\documentclass[runningheads]{llncs}

\usepackage{eccv}

\usepackage{eccvabbrv}

\usepackage{graphicx}
\usepackage{booktabs}
\usepackage{esvect}
\usepackage[accsupp]{axessibility} 
\usepackage[export]{adjustbox}

\usepackage{multirow}
\usepackage{amsmath}
\usepackage{dsfont}

\makeatletter
\newcommand*{\overrightharpoonup}{\mathpalette{\overarrow@\rightharpoonupfill@}}
\newcommand*{\rightharpoonupfill@}{\arrowfill@\relbar\relbar\rightharpoonup}
\makeatother

\usepackage{hyperref}
\usepackage{url}
\usepackage{orcidlink}

\DeclareMathOperator{\vectorprojection}{proj}
\NewDocumentCommand{\proj}{s e{_} m}{%
    \IfBooleanTF{#1}{#3_{\stretchrel*{\parallel}{\perp}#2}}{\vectorprojection_{#2}#3}
}
\newcommand{\norme}[2][]{\ensuremath{\left\lVert #2 \right\rVert_{#1}}}

\begin{document}

\title{Equi-GSPR: Equivariant SE(3) Graph Network Model for Sparse Point Cloud Registration} 


\titlerunning{Equi-GSPR}

\author{Xueyang Kang\inst{1, 2, 3}\orcidlink{0000-0001-7159-676X} \and
Zhaoliang Luan\inst{2, 4}\orcidlink{0000-0002-9345-5455} 
\and
Kourosh Khoshelham\inst{3}\orcidlink{0000-0001-6639-1727}
\and
\\
Bing Wang\inst{2}\orcidlink{0000-0003-0977-0426}\thanks{Corresponding author}}

\authorrunning{X. Kang, Z. Luan, K. Khoshelham and B. Wang$^{\star}$}

\institute{Faculty of Electrical Engineering, KU Leuven \and Spatial Intelligence Group, The Hong Kong Polytechnic University \and Faculty of Engineering and IT, The University of Melbourne \and IoTUS Lab, Queen Mary University of London
\\
\email{alex.kang@kuleuven.com, z.luan@qmul.ac.uk, k.khoshelham@unimelb.edu.au, bingwang@polyu.edu.hk}
}

\maketitle

\begin{figure}[!h]
\vspace{-3.6em}
  \centering
  \includegraphics[width=0.86\textwidth]{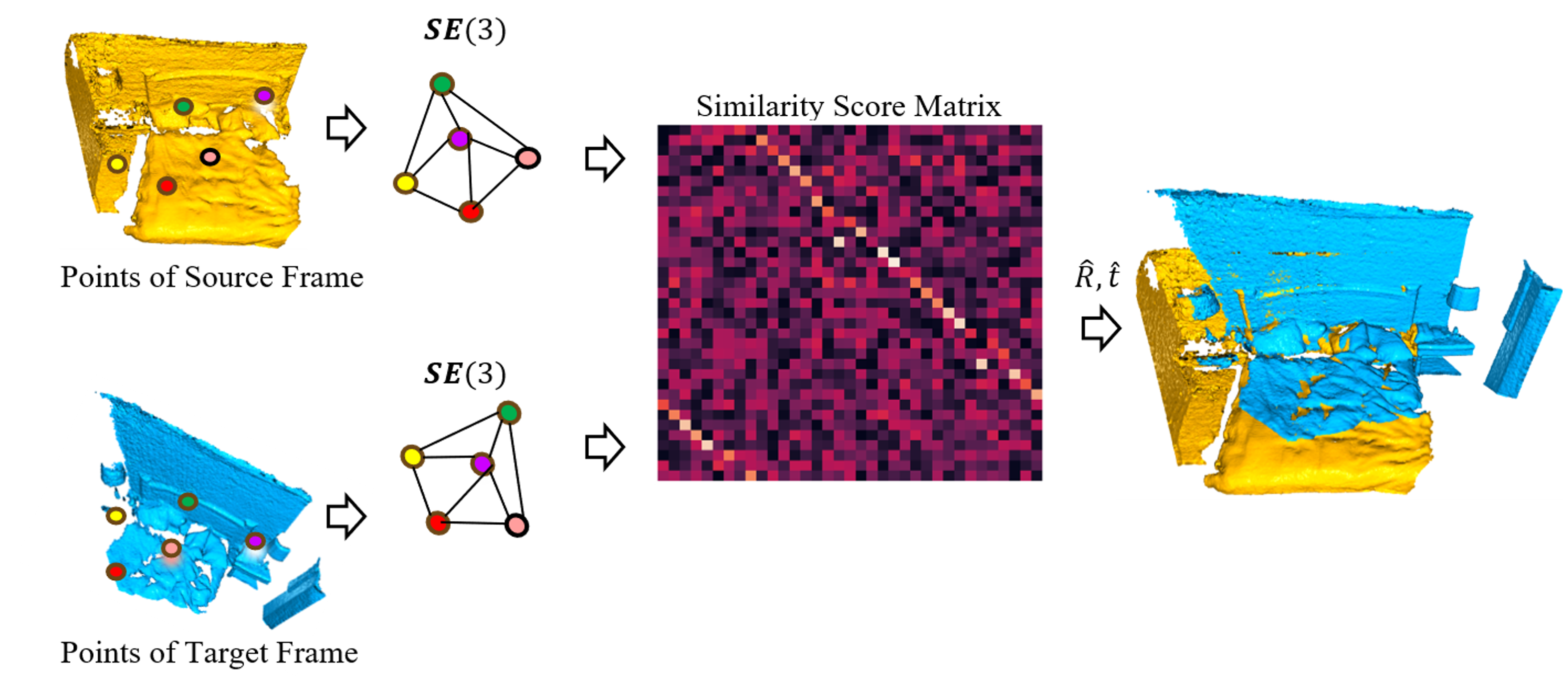}
  \vspace{-.9em}
  \caption{The registration model converts the sparse point descriptors of the source and target frames into an equivariant graph feature representation, respectively. Then the $\mathbf{SE}(3)$ equivariant graph features are used for the similarity score calculation. The matched features are then decoded into the relative transform to align the two scans. 
 }
 \label{fig:teaser}
\vspace{-4.5em}
\end{figure}

\begin{abstract} 
Point cloud registration is a foundational task for 3D alignment and reconstruction applications. While both traditional and learning-based registration approaches have succeeded, leveraging the intrinsic symmetry of point cloud data, including rotation equivariance, has received insufficient attention. This prohibits the model from learning effectively, resulting in a requirement for more training data and increased model complexity. To address these challenges, we propose a graph neural network model embedded with a local Spherical Euclidean 3D equivariance property through $\mathbf{SE}$(3) message passing based propagation. Our model is composed mainly of a descriptor module, equivariant graph layers, match similarity, and the final regression layers. Such modular design enables us to utilize sparsely sampled input points and initialize the descriptor by self-trained or pre-trained geometric feature descriptors easily. Experiments conducted on the 3DMatch and KITTI datasets exhibit the compelling and robust performance of our model compared to state-of-the-art approaches, while the model complexity remains relatively low at the same time. Implementation code can be found here, \url{https://github.com/alexandor91/se3-equi-graph-registration}.
\end{abstract}
\section{Introduction}
\label{sec:intro}
The registration of point clouds typically involves formulating robust geometric feature descriptors and a subsequent complex matching process to predict feature correspondences \cite{cheng2023sampling}. However, these correspondences established from raw point cloud often exhibit a high outlier-to-inlier ratio, leading to significant registration errors or complete failures. To enhance the robustness of registration processes, PointDSC \cite{bai2021pointdsc} explicitly calculates local feature spatial consistency and evaluates pairwise 3D geometric feature descriptor similarity across two frames \cite{choy2019fully, wang2019deep} to eliminate outliers from the alignment optimization process. Other approaches like Deep Global Registration (DGR) \cite{choy2020deep} treat correspondence prediction as a classification problem, utilizing concatenated coordinates of input point cloud pairs and employing a differentiable optimizer for pose refinement. Despite the effectiveness of these models on public datasets, their training requires accurate correspondence supervision, necessitating a complex point-to-point search process that is particularly vulnerable to numerous outliers.

Geometric feature descriptors, derived from keypoint neighborhoods of a specified range, often overlook the underlying geometric topology of the data, such as the global connectivity among points. This oversight results in feature descriptors lacking $\mathbf{SE}$(3) rotation equivariance, thereby impeding efficient and robust learning of rotation-equivariant and invariant features. The recently introduced RoReg \cite{wang2023roreg} model employs a rotation-guided detector to enhance rotation coherence matching and integrates it with RANSAC for pose estimation. However, it suffers from high computational demand and reduced processing speed. This highlights the need for more efficient rotation-equivariant model architectures to significantly enhance registration performance.

To address these challenges, we introduce a novel approach that leverages a graph convolution-based model to jointly learn $\mathbf{SE}$(3) equivariant features, starting with feature descriptors extracted from sparsely sampled points across two frames. Our proposed $\mathbf{SE}$(3) equivariant graph network model, aimed at sparse point cloud registration, is depicted in \cref{fig:teaser}. Unlike Transformer and CNN-based models, our graph architecture captures both the topology and geometric features of point clouds, similar to other proposed geometric descriptors \cite{shi2020point, zhang2018graph}, facilitating the learning of fine-grained rigid rotation-equivariant feature representations for more robust and coherent point cloud registration through data symmetry. The primary contributions of our study are the following:
\begin{itemize}
\item Introduction of an equivariant graph model to facilitate neighbor feature aggregation and $\mathbf{SE}$(3) equivariant coordinate embedding from either learned geometric descriptors for point cloud registration.
\item Implementation of a novel matching approach within the implicit feature space, based on similarity evaluation and Low-Rank Feature Transformation (LRFT), eliminating the need for explicit point correspondence supervision and exhausting search.
\item Development of a specific matrix rank-based regularizer to enable the model to automatically identify and mitigate the impact of correspondence outliers, enhancing the robustness of the registration process.
\end{itemize}

\section{Related Work}
The concept of equivariant properties can be embedded in the layers of Convolutional Neural Networks (CNNs) to depict $\mathbf{SO}$(2) group characteristics, as initially proposed by Cohen \cite{cohen2016group}, and later extended to encompass arbitrary continuous input \cite{finzi2020generalizing}. Another area of study focuses on equivariance representation by employing steerable kernel filters \cite{cohen2018spherical, weiler2018learning, weiler20183d, sosnovik2019scale, e2cnn, jenner2022steerable} for equivariance learning. For more intricate tasks involving $\mathbf{SO}$(3), techniques such as Vector Neurons \cite{deng2021vector} and Tensor Field Networks \cite{thomas2018tensor} can be viewed as implementations of the capsule network model \cite{xinyi2018capsule}, transitioning from scalar values to vectors. Following the introduction of the Transformer model, Lie-group-based Transformer models \cite{hutchinson2020lietransformer, fuchs2020se, fuchs2021iterative} have been developed to capture equivariance through attention mechanisms. Moreover, to address the intricate equivariance inherent in the input data, equivariant($n$) graph neural networks \cite{keriven2019universal, satorras2021en, du2022se} are utilized to learn equivariant features for dynamic and complex issues. As $\mathbf{SE}$(3) features are maintained through message passing \cite{brandstetter2021geometric} within the graph model, these equivariant models exhibit considerable potential in addressing many longstanding challenges, such as predicting molecule structures \cite{schutt2018schnet} or quantum structures \cite{gilmer2017neural}, as well as particle dynamic flow physics \cite{kohler2020equivariant, bogatskiy2020lorentz}. Some studies have explored the application of equivariant models in various point cloud tasks, including 3D detection \cite{shi2020point}, 3D point classification \cite{zhang2018graph}, point cloud-based place recognition \cite{lin2023se}, 3D shape point registration \cite{chen2021equivariant, zhu2023e2pn}, and 3D shape reconstruction \cite{chatzipantazis2022se}. These applications underscore the learning efficiency gained from the leveraging of the intrinsic symmetries in input data.

Despite the prevalence of equivariant models in the microscopic realm \cite{schutt2018schnet, fuchs2020se}, the application of such models for tasks like multi-view 3D reconstruction \cite{gojcic2020learning} or other intricate 3D challenges remains under-explored. A fundamental component of 3D reconstruction involves point cloud registration. Many conventional methods employ linear-algebra-based optimization techniques for iterative point cloud registration, such as point-to-plane registration \cite{park2003accurate, park2017colored}, LOAM \cite{wang2021f}, and its variation F-LOAM \cite{wang2021f}. In recent years, there has been a rise in deep learning models for registration purposes, relying on representative feature descriptors \cite{wang2019deep, choy2019fully} or precise correspondence establishment between descriptors, like deep global registration \cite{choy2020deep}. To enhance registration accuracy, some studies focus on developing more resilient descriptors like rotation-equivariant descriptors \cite{wang2022you, wang2023roreg, bai2020d3feat} for subsequent correspondence matching or incorporating $\mathbf{SO}$(2) rotation equivariance into the registration framework using cylindrical convolution, as in Spinnet \cite{ao2021spinnet}. Other research works concentrate on optimizing correspondence search explicitly, for instance, Stickypillars utilizes optimal transport for matching, and PointDSC \cite{bai2021pointdsc} employs spectral matching to eliminate outliers from raw correspondences. In contrast to conventional models, works in \cite{zhou2016fast, qin2022geometric} excel in rapid registration performance. Moreover, some end-to-end models improve overall performance from feature descriptor learning to feature association in a differentiable way, like 3D RegNet \cite{pais20203dregnet}, or correspondence-free registration aimed at streamlining the 3D point cloud registration \cite{zhu2022correspondence}. Notably, Banani et al. \cite{el2021unsupervisedr} introduce an unsupervised model for point cloud registration using differentiable rendering, while Predator \cite{Huang_2021_CVPR} demonstrates registration capabilities even in applications with low input point cloud overlap and numerous outliers.
\section{Method}
\begin{figure}[!tbp]
  \centering
  \includegraphics[width=0.96\textwidth]{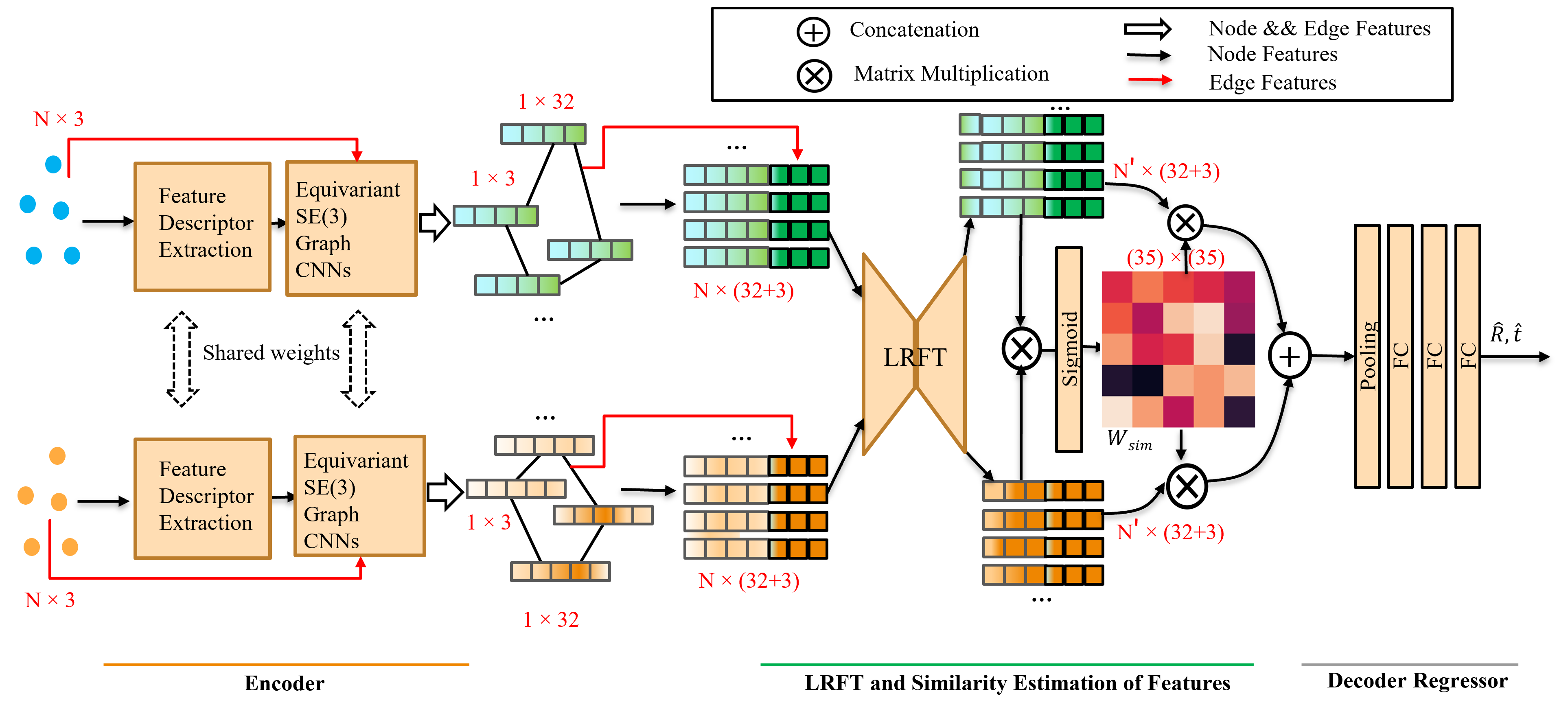}
  \caption{The registration model consists of an encoder, a feature match block, and a decoder. Pointwise feature descriptors are extracted from the source and target scan points, passed through equivariant graph layers, and combined with coordinate embeddings to form a row-major order matrix. Next, the feature matrices from the source and target frames are compressed using MLPs-based Low-Rank Feature Transformation (LRFT). The aggregated features are used to create a similarity map through dot product of feature descriptors. In the decoder, features are weighted by similarity scores, then concatenated, and processed through pooling and fully connected layers to predict relative translation $t_j^i$ and quaternion $q_j^i$. 
 }
 \label{fig:overview}
\vspace{-2.0em}
\end{figure}

Our registration process begins by extracting feature descriptors from downsampled point clouds. Equivariance is integrated into the features through equivariant graph convolution layers. Subsequently, the number of features in the pairwise graph is aggregated to a smaller number via Low-Rank based constraint. Finally, the similarity between the pairwise features of the two frames is calculated for relative transform prediction. A detailed illustration of the model is shown in \cref{fig:overview}. The input to our model consists of $N$ points $\Vec{X}=[\Vec{x}_1, \ldots, \Vec{x}_N]$ $\in N\times\mathbb{R}^{3}$ from the source frame, and $N$ points $\Vec{Y}=[\Vec{y}_1, \ldots, \Vec{y}_N]$ $\in N\times\mathbb{R}^{3}$ from the target frame, where $x_i\in\mathbb{R}^{3}$ and $y_j\in\mathbb{R}^{3}$ form a correspondence $(i, j)$. It is important to note, for ease of subsequent similarity search, that the coordinates of points in each frame are rearranged in descending order based on the ray length $||\mathbf{r}(t)||^2$ from the point position to the sensor frame center $\mathbf{o_{s}}$. For numerical stability during training, the source scan is normalized to a canonical frame, and the target scan is transformed relative to the source frame, allowing the model to predict the relative transformation from source to target.
\subsection{Feature Descriptor}
We incorporate geometric details of nearby interest points into our graph model using feature descriptors. To extract these descriptors, we can reuse available pre-trained point-based descriptors or train a shallow Multi-Layer Perceptron (MLP) module with $l_1$ layers prior to the model in an end-to-end manner, inspired by PointNet++ \cite{qi2017pointnet++}. The feature representation of point $i$ in the next layer $l+1$ is calculated by averaging the output of the mapping function $h(\cdot)$ applied to the relative positional coordinates of point $i$ and neighboring point $k\in \mathcal{N}(i)$ ($n$ points), along with the hidden feature $h_{k}^{l_1}$ from the prior layer $l_1$.
\begin{align}
    \Vec{h}_i^{l_1+1} &= \frac{1}{n_{}}\sum_{k\in \mathcal{N}(i)} f_{h}(\Vec{h}_k^{l_1}, \Vec{x}_{k}-\Vec{x}_{i}).
    \label{eq:pointnet-mlp}
\end{align}
\subsection{Equivariant Graph Network Model}
By utilizing the equivariant graph representation ($l_2$ layers) as introduced by Satorras et al. \cite{satorras2021en}, we can enhance the receptive field and representation for feature descriptors of interest points by incorporating $\mathbf{SE}(3)$ equivariant properties via graph feature aggregation. Our method leverages graph-based message passing techniques \cite{gilmer2020message} to propagate $\mathbf{SE}(3)$ equi-features. For the construction of the graph $\mathcal{G} = (\mathcal{V}, \mathcal{E})$ with vertices $\mathcal{V}$ and $\mathcal{E}$ edges. The individual hidden point descriptors $\Vec{h}_i^{l_2}\in \mathbb{R}^{32}$ and the point coordinate embedding $\Vec{x}_i^{l_2}\in \mathbb{R}^{3}$ at layer $l_2$ are treated as the node and edge features, respectively. The graph convolutional layer updates the edge equi-message $\Vec{m}_{ik}\in \mathbb{R}^{3\times 3}$, node hidden feature $\Vec{h}_i^{l_2} \in \mathbb{R}^{32}$, and coordinate embedding $\Vec{x}_i^{l_2}\in \mathbb{R}^{3}$ at each equivariant layer.
\begin{align}
    \Vec{m}_{ik} &= \phi_{m}(\Vec{h}_{i}^{l_2}, \Vec{h}_{k}^{l_2}, \norme{\Vec{x}_k^{l_2} - \Vec{x}_i^{l_2}}^\frac{1}{2}), \label{eq:m}\\
    \Vec{x}_i^{l_2+1} &= \Vec{x}_i^{l_2}+C\sum_{k\in \mathcal{N}(i)}\exp(\Vec{x}_k^{l_2} - \Vec{x}_i^{l_2})\phi_x(\proj_{\Vec{\mathcal{F}}_{ik}}{\Vec{m}_{ik}}), \label{eq:x}\\
    \Vec{h}_i^{l_2+1} &= \phi_{h}(\Vec{h}_i^{l_2}, \sum_{k\in \mathcal{N}(i)} (\proj_{\Vec{\mathcal{F}}_{ik}}\Vec{m}_{ik})), \label{eq:h}
\end{align}
where $\phi_{m}$, $\phi_{x}$, and $\phi_{h}$ represent 1D convolutional layers for the message, coordinate embedding, and hidden feature update, respectively. The normalizing factor $C$ is applied to the exponentially weighted sum of mapped equi-message in \cref{eq:x}. Additionally, a neighbouring search of $\Vec{x}_i^{l_2}$ within a specific radius is conducted to find the $\mathcal{N}(i)$ neighbouring feature descriptors for edge establishments, and this is used to prevent information overflow by confining the exchange of information within a local context, thereby reducing the complexity of the graph feature adjacency matrix from $O(n^2)$ to approximately $O(n)$. In \cref{eq:x}, the projection of $\Vec{m}_{ij}$ onto a locally equivariant frame ($\proj_{\Vec{\mathcal{F}}_{ik}}(\cdot)$) helps to preserve the $\mathbf{SO}(3)$ feature invariance. The frame $\Vec{\mathcal{F}}_{ik}$ is constructed using pairwise coordinate embeddings as outlined in ClofNet \cite{du2022se},
\begin{align}
    \Vec{\mathcal{F}}_{ik} &= (\Vec{a}_{ik}, \Vec{b}_{ik}, \Vec{c}_{ik}), \label{eq:F_{ij}} \\
    &= (\frac{\Vec{x}_i^l - \Vec{x}_k^l}{\|\Vec{x}_i^l - \Vec{x}_k^l\|}, \frac{\Vec{x}_i^l \times \Vec{x}_k^l}{\|\Vec{x}_i^l \times \Vec{x}_k^l\|}, \frac{\Vec{x}_i^l - \Vec{x}_k^l}{\|\Vec{x}_i^l - \Vec{x}_k^l\|} \times \frac{\Vec{x}_i^l \times \Vec{x}_k^l}{\|\Vec{x}_i^l \times \Vec{x}_k^l\|}), \label{eq:equi}
\end{align}
Consequently, the projection of $\Vec{m}_{ik}$ into $\hat{\Vec{m}}_{ik}$ is formulated into the linear combination of axes of the local equi-frame scaled by the coefficients ($x_{ik}^{\Vec{a}}, x_{ik}^{\Vec{b}}, x_{ik}^{\Vec{b}}$),
\begin{align}
\vspace{-1.8em}
   \proj_{\Vec{\mathcal{F}}_{ik}}\Vec{m}_{ik} =\hat{\Vec{m}}_{ik} = x_{ik}^{\Vec{a}} \Vec{a}_{ik} + x_{ik}^{\Vec{b}} \Vec{b}_{ik} + x_{ik}^{\Vec{c}} \Vec{c}_{ik}.
    \label{eq:m_{ij}}
\end{align}
The projection of edge message $m_{ij}$ in \cref{eq:x} is performed in the local equi-frame (within bracket of \cref{eq:equi}), to obtain projected message $\hat{\Vec{m}}_{ik}$, while the scalar coefficients in \cref{eq:m_{ij}} remain $\mathbf{SO}(3)$ invariant. Consequently, the sum of equi-projected message in \cref{eq:h} is still a vector-based sum, maintaining the equivariance upon integration into the hidden layer $\phi_{h}$.
\subsection{Low-Rank Feature Transformation}
\begin{figure}[!ht]   
\vspace{-1.5em}
\begin{subfigure}{0.36\textwidth}
  \includegraphics[width=\textwidth, height=6.0cm]{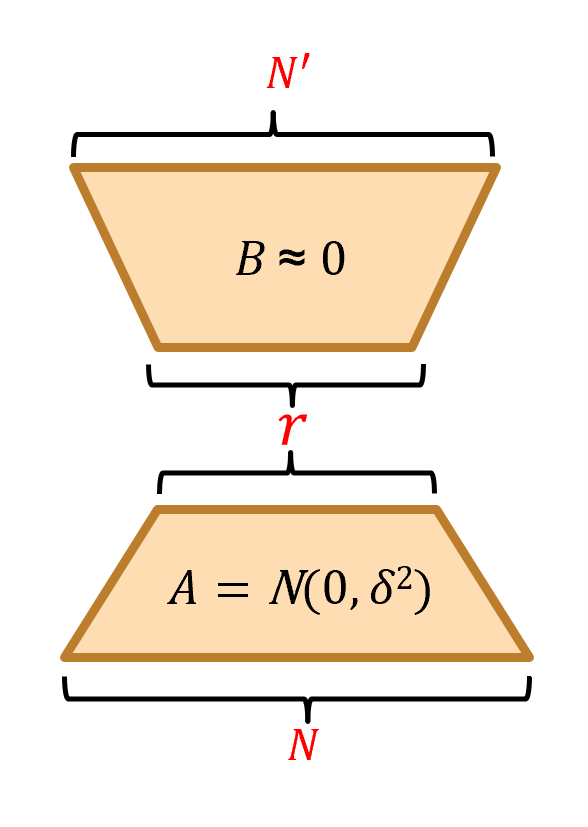}
    \caption{Feature number reduction by Low-Rank Feature Transformation.}
    \vspace{-0.0em}
    \label{fig:LRFT}
\end{subfigure}
\hfill
\begin{subfigure}{0.58\textwidth}
    \includegraphics[width=\textwidth, height=5.6cm]{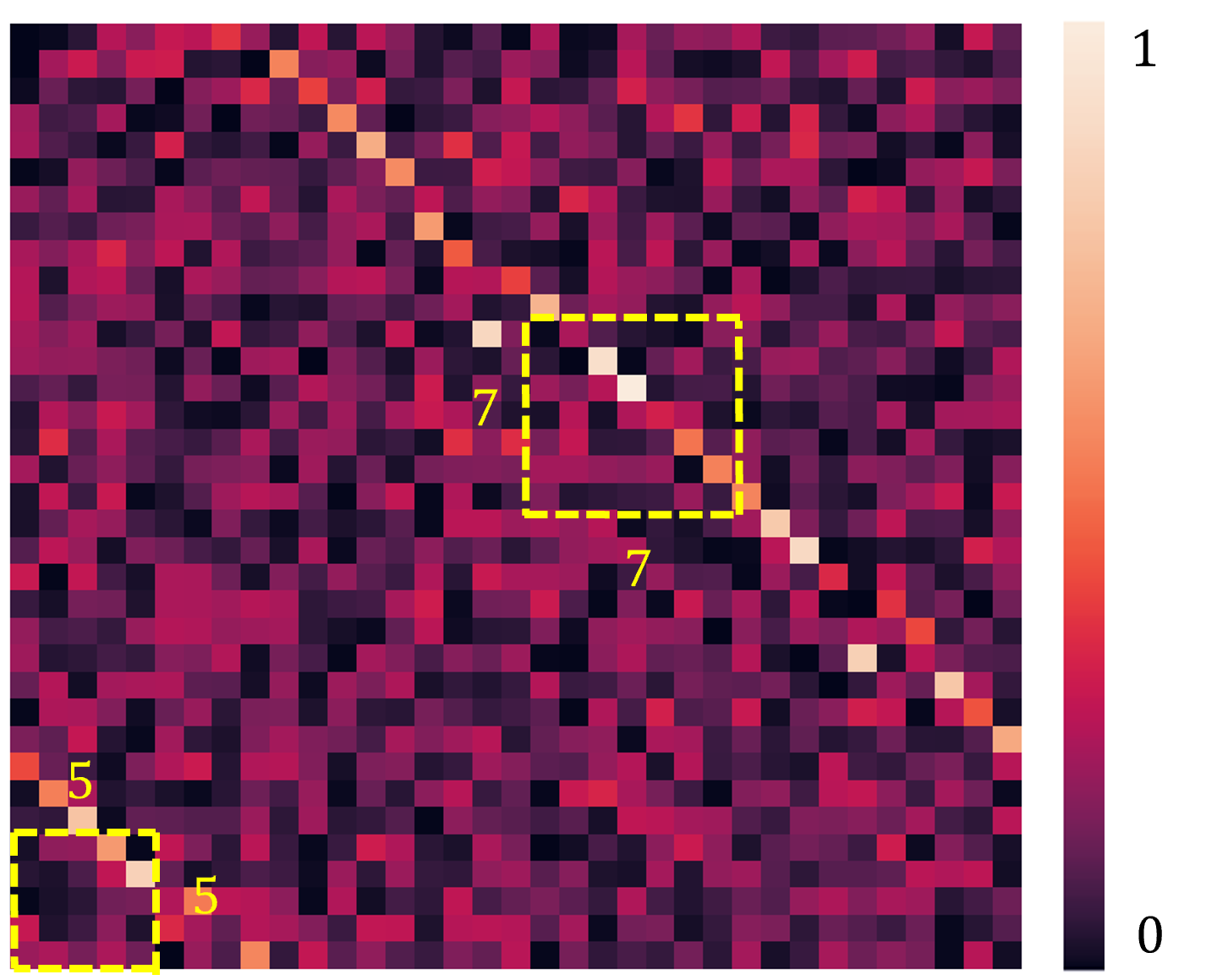}
    \caption{Zoomed-in submatrix ($35\times35$) from full similarity score matrix $\Vec{S}$ ($N'\times N'$).}
    \vspace{-0.0em}
    \label{fig:sim}
\end{subfigure}
\caption{Reducing the feature number through low-rank using MLP layers (a), and examining the similarity score matrix with submatrices for rank verification at bottom right ($5\times 5$) and center ($7\times 7$) of yellow dashed region as illustrated in subfigure (b).}
\label{fig:dim-reduc}
\vspace{-1.3em}
\end{figure}

Inspired by LoRA of language model \cite{hu2021lora}, our approach diverges by not requiring pre-trained weights for fine-tuning. We employ two stacked linear forward layers with low-rank constraints in the middle of model (\cref{fig:LRFT}) to map feature descriptors to aggregated descriptors. We name it as Low-Rank Feature Transformation (LRFT). This design enhances similarity match reliability and computational efficiency by performing matches on aggregated descriptors with integrated neighboring information. Specifically, The motivation for employing LRFT is twofold: 1) Theoretically, low-rank constraints in linear layers capture essential feature correlations within descriptors, as demonstrated by the matrix low-rank theorem (refer to Appendix Sec.1.2), leading to more reliable similarity matches; 2) Practically, our LRFT improves computational efficiency by aggregating feature descriptors prior to similarity matching, and It also enhances low-rank learning efficiency by training parameters during the forward pass, eliminating the need for pre-trained weights.

After the final layer (${(l_2^*)}$th) of the equivariant graph module, the output graph features consist of both node and edge features. During this stage, we preserve each graph node feature $\Vec{h}_i^{l_2^*}, i\in N$ from the source frame and feature $\Vec{h}_j^{l_2^*}, j\in N$ from the target frame of the last graph layer. Next, the node feature is combined with the mean coordinate embeddings $\hat{\Vec{x}}_i^{l_2^*}  = \frac{1}{n}\sum_{k\in\mathcal{N}(i)} \Vec{x}_k^{l_2^*}$, where $ \Vec{x}_k^{l_2^*}\in\mathbb{R}^3$ is obtained from the edge embeddings (\cref{eq:x}) connected to the node feature $\Vec{x}_i^{l_2^*}$. Consequently, matrices for the source frame $\Vec{H}_{src} \in \mathbb{R}^{N\times 35}$ and target frame $\Vec{H}_{tar}\in \mathbb{R}^{N\times 35}$ are created by stacking of node features $(\Vec{h}_i^{l_2^*}, \hat{\Vec{x}}_i^{l_2^*})$ and coordinate embeddings along the column respectively. Before computing the feature similarity, the Low-Rank Feature Transformation (LRFT) technique is applied to compress the features into $\hat{\Vec{H}}_{src}\in \mathbb{R}^{N'\times 35}$ and $\hat{\Vec{H}}_{tar}\in\mathbb{R}^{N'\times 35}$ utilizing parameters of $\Vec{A}$ and $\Vec{B}$ mapping layers.
\begin{align}
   \hat{\Vec{H}}_{src}, \hat{\Vec{H}}_{tar} = (\Vec{A}\Vec{B})^{\Vec{T}}(\Vec{H}_{src}, \Vec{H}_{tar}),  \label{eq:H}
\end{align}
where $\Vec{A}$ is a $N\times r$ matrix and $\Vec{B}$ is $r\times N'$ matrix, and rank $r\ll\min(N, N')$, so that the number of node features are compressed into $N'$ dimension after LRFT module mapping via multiplication $\Vec{A}\Vec{B}$, as depicted in the left of \cref{fig:LRFT}. $\Vec{A}$ is initialized from Gaussian distribution with standard deviation $\delta = \sqrt{r}$, while $\Vec{B}$ is initialized with small constant close to zero. 

The LRFT layers extract spatial context from neighboring feature descriptors through linear mapping, efficiently aggregating local information for decoder.
\subsection{Similarity Calculation}
Subsequently, we calculate the feature similarity score matrix (refer to \cref{fig:sim}) for feature correspondence establishments using the compressed number of features after the LRFT module. This is achieved by computing the dot product $<\cdot>$ of features as an element. Prior to the multiplication, the respective element features $\Vec{h}_i$ from $\hat{\Vec{H}}_{src}\in\mathbb{R}^{N'\times 35}$ and $\Vec{h}_j$ from $\hat{\Vec{H}}_{tar}\in\mathbb{R}^{N'\times 35}$ are first normalized to $\Vec{\hat{h}}_i$ and $\Vec{\hat{h}}_j$.
\begin{align}
    \Vec{S}_{ij} = <\Vec{\hat{h}}_i\cdot\Vec{\hat{h}}_j>,  \label{eq:s_{ij}}
\end{align}
which forms a square similarity matrix $\Vec{S}\in\mathbb{R}^{N'\times N'}$, then subsequently normalized along each row to produce $\hat{\Vec{S}}\in\mathbb{R}^{N'\times N'}$. We compute the determinant of $\hat{\Vec{S}}$ by calculating trace to indicate whether the matrix rank has deficiency, which may arise from the presence of ambiguous correspondences. Accordingly, as per the match assignment rule, each row of the similarity matrix $\hat{\Vec{S}}$ should contain a singular value close to one, depicted as a light-colored square in \cref{fig:sim}. Subsequently, the similarity matrix is employed to project the feature matrices $\hat{\Vec{H}}_{src}$ and $\hat{\Vec{H}}_{tar}$ through multiplication by $\hat{\Vec{S}}_{}^{T}\hat{\Vec{H}}_{src}\in \mathbb{R}^{N'\times 35}$ and $\hat{\Vec{S}_{}}\hat{\Vec{H}}_{tar}\in \mathbb{R}^{N'\times 35}$, respectively. The resulting matrices are concatenated to facilitate subsequent pooling and mapping through fully-connected layers.
Furthermore, a regularizer is used to enforce the rank of $\hat{\Vec{S}}$ close to $r$, ensuring that a submatrix $\hat{\Vec{S'}}$ with rank $r$ out of $\hat{\Vec{S}}$ can be found,
\begin{align}
    \vspace{-1.5cm}
    \mathcal{L}_{Reg} = \left|(\textbf{Trace}(\hat{\Vec{S}}^{T}\hat{\Vec{S}}))^{\frac{1}{2}} - r \right|.   \label{eq:Reg}
\end{align}
Additionally, to eliminate outlier feature correspondences, each matched pair element $\hat{\Vec{S}}_{ij}$ undergoes verification through a submatrix full-rank check. This involves evaluating the determinant of a $7\times7$ submatrix centered at the feature element $\hat{\Vec{S}}_{ij}$ or a $5\times5$ submatrix at the border element of the $\hat{\Vec{S}}$ matrix (highlighted by the red dashed box in \cref{fig:sim}). This verification process ensures local consistency in feature similarity matches, aiding in the identification of globally consistent and reliable match pattern search from the similarity matrix. Following verification, the final valid rank of the similarity matrix $\hat{\Vec{S}}$ is established as $r \leq 128$, with any invalid assignment row $\hat{\Vec{S}}_{i\cdot}$ zeroed out to mask the corresponding feature for subsequent computations. The upper limit rank $r$ is derived from the \emph{\textbf{Theorem}}: $\mathbf{Rank(\Vec{A}\Vec{B}) \leq \min(Rank(\Vec{A}), Rank(\Vec{B}))}$. Please refer to the supplementary part for detailed proof the rank theorem. 

\subsection{Training Loss}
The final layer predicts translation $\hat{\Vec{t}}$ and rotation matrix $\hat{\Vec{R}}$ in quaternion form. The ground-truth translation and rotation are denoted by $\Vec{t}^{*}$ and $\Vec{R}^{*}$ respectively. 
\begin{equation}
 \mathcal{L}_{total}=  \mathcal{L}_{rot} + \mathcal{L}_{trans} + \beta \mathcal{L}_{Reg},
 \label{eq:total_loss}
\end{equation}       
$\beta$ for regularizer is set to 0.05. The translation error (TE) and rotation error (RE) losses can be formulated as follows,
\begin{align}
    \mathcal{L}_{rot}(\hat{R}) &= \arccos{\frac{\textbf{Trace}(\hat{\Vec{R}}^T{\Vec{R}}^{*})-1}{2}}, \\
    \mathcal{L}_{trans}(\hat{t}) &= {\|\hat{\Vec{t}}- {\Vec{t}}^{*}\|}^2.
    \label{eq:tran-r-loss}
\end{align}
The rotation error term $\mathcal{L}_{rot}$ and translation error term $\mathcal{L}_{trans}$ are measured in radians and meters, respectively. Given that the predicted transform is relative, from source to target frame, the scale of these transforms is typically in normal scale to avoid numerical stability issues in training.
\section{Experiments}
We evaluate the performance of the proposed model for point cloud registration in both indoor and outdoor environments. For indoor scenes, we utilize 3DMatch introduced by Zeng et al. \cite{zeng20173dmatch}. The raw point clouds are uniformly downsampled to 1024 points through a voxel filter. For the outdoor evaluation, we select the KITTI dataset \cite{geiger2012we} with the same dataset split from the creators and follow the same downsampling process as in Choy \etal, \cite{choy2019fully}. We report both the qualitative and quantitative results of the proposed model. Furthermore, we offer an in-depth analysis of the effect of varying parameter configurations and the contribution of each component to enhancing the model's performance. The computational efficiency of each model is presented in the metric table below.\\

\noindent \textbf{Implementation Details.} The key parameters of our model include the dimension of graph-relevant features and LRFT layers. Initially, the extracted feature descriptor dimension is 32 for subsequent graph learning. A critical aspect is the number of nearest neighbors for each node feature in the graph, set to 16 for constructing the graph using ball query for 3DMatch (ball radius at 0.3$m$), while for KITTI, we employ kNN, selecting the nearest 16 points of query point to form a graph with $1024$ nodes and $1024 \times 16$ edges. In graph learning, the node feature dimension is $32$, and the edge embedding feature is $3$, comprising coefficients projected onto the locally constructed coordinate frame as shown in \cref{eq:m_{ij}}. We use $4$ equi-graph layers throughout the tests. The LRFT module consists of $3$ parameters: input dimension $N$, internal rank $r$, and $N'$. Our model adopts a configuration of $1024/(32+3)/128$, where rank $r$ is the sum of graph node feature dimension ($32$) and coordinate embedding dimension 3. The submatrix determinant check for similarity score matrix $\hat{\Vec{S}}$ is $5\times5$ along the borders and $7\times7$ within the matrix. A performance analysis comparing different parameter configurations is presented in the subsequent ablation section. All training and inference tasks are conducted on a single RTX 3090 GPU.\\  

\noindent \textbf{Evaluation Metrics.} We employ the average Relative Error (RE) and Translation Error (TE) metrics from PointDSC \cite{bai2021pointdsc} to assess the accuracy of predicted pose errors in successful registration. Additionally, we incorporate Registration Recall (RR) and $\emph{F1 score}$ as performance evaluation measures. To evaluate these metrics, we establish potential corresponding point pairs $(\Vec{x}_i, \Vec{y}_j) \in \Omega$ using input points from two frames, following the correspondence establishment approach outlined in PointDSC \cite{bai2021pointdsc}. We apply the predicted transformation to the source frame point $\Vec{x}_{i}^{}$, recording a pairwise registration success only when the average Root Mean Square Error (RMSE) falls below a predefined threshold $\tau$. The registration recall value $\delta{}$ is calculated as:
\begin{align}
    \delta_{} &= \sqrt{\frac{1}{\mathcal{N}(\Omega)}\sum_{(\Vec{x}_{i}, \Vec{y}_{j})\in\Omega}\mathds{1}[\|\hat{\Vec{R}} \Vec{x}_{i} + \hat{\Vec{t}} - \Vec{y}_{j} \|^2 < \tau]},
    \label{eq:rmse-rr}
\end{align}
where $\mathcal{N}(\Omega)$ denotes the total number of ground truth correspondences in set $\Omega$. The symbol $\mathds{1}$ functions as an indicator for condition satisfaction. Removing the conditional check within the indicator brackets on the equation's right side transforms it into a standard Root Mean Square Error (RMSE), 

$\sqrt{\frac{1}{\mathcal{N}(\Omega)}\sum_{(\Vec{x}{i}, \Vec{y}{j})\in\Omega}|\hat{\Vec{R}}\Vec{x}{i} + \hat{\Vec{t}}-\Vec{y}{j} |^2}$. This RMSE metric is utilized in ablation experiments for parameter analysis. The $\emph{F1 score}$ is defined as $2\cdot \frac{Precision\times Recall}{Precision + Recall}$.\\  

\noindent \textbf{Baseline Methods} 1) For the 3DMatch \cite{zeng20173dmatch} benchmark, we compare our model with vanilla RANSAC implementations using various iterations and optimization refinements. We also include Go-ICP \cite{yang2015go} and Super4PCS \cite{mellado2014super}, which operate on raw points. Among learning-based methods, we compare with DGR \cite{choy2020deep} and PointDSC \cite{bai2021pointdsc} combined with FCGF descriptors \cite{choy2019fully}. Additionally, we select D3Feat \cite{bai2020d3feat}, SpinNet \cite{ao2021spinnet}, and RoReg \cite{wang2023roreg}, which incorporate rotation invariance or equivariance. These learning methods do not support descriptor replacement, denoted by $^*$. 2) For KITTI sequences \cite{geiger2012we}, we implement the hand-crafted feature descriptor FPHF \cite{rusu2009fast} due to performance saturation issues with FCGF \cite{choy2019fully} descriptors, as noted in PointDSC \cite{bai2021pointdsc}. RoReg \cite{wang2023roreg} is replaced with the registration model from the FCGF paper \cite{choy2019fully} (denoted as FCGF-Reg) due to public code limitations for KITTI dataset.
\subsection{Indoor Fragments/Scans Registration}
Point clouds are initially downsampled using a 5cm voxel size to generate 1024 sampled points. Registration success is evaluated using thresholds of 30cm for translational error (TE) and $15^{\circ}$ for rotational error (RE). The correspondence distance threshold $\tau$ in \cref{eq:rmse-rr} is set at 10cm. Comparative results between our proposed model and baseline approaches are presented in \cref{tab:cmps-3dmatch}. Our model outperforms all comparison methods, despite slightly slower latency compared to RANSAC with 1k iterations. RoReg and SpinNet, ranking second and third, demonstrate minimal registration errors and maximal registration scores, highlighting the advantages of incorporating rotation features. While our model can integrate the FCGF descriptor, we present evaluation results using the PointNet++ learning descriptor for end-to-end training.
\begin{table}[!th]
\centering
    \caption{Evaluation results of registration methods on 3DMatch show non-learning-based approaches at the top and deep-learning registration models below. The learning-based feature descriptor FCGF \cite{choy2019fully} is tested with various learning baseline approaches. The symbol $^*$ indicates the original model implementation on 3D Match due to the lack of support for this feature descriptor replacement.}
    \label{tab:cmps-3dmatch}
\begin{adjustbox}{width=0.85\textwidth}    
     \begin{tabular}{c c c c c c } 
        \toprule 
        & RE($^{\circ}$) $\downarrow$ & TE(cm) $\downarrow$ & RR(\%) $\uparrow$ & F1(\%) $\uparrow$  & Time(s) $\downarrow$ \\ 
       \hline
       RANSAC-1k \cite{fischler1981random} & 3.16 & 9.67 &  86.57 &  76.62 & $\mathbf{0.08}$ \\ 
       \hline
       RANSAC-10k & 2.69 & 8.25 & 90.70 & 80.76 & 0.58 \\ 
       \hline
       RANSAC-100k & 2.49 & 7.54 & 91.50 & 81.43 & 5.50 \\ 
       \hline
       RANSAC-100k + refine & 2.17 & 6.76 &  92.30 &  81.43 & 5.51 \\ 
       \hline
        Go-ICP \cite{yang2015go} & 5.38 & 14.70 & 22.95 & 20.08 & 771.0  \\ 
        \hline
        Super4PCS \cite{mellado2014super} & 5.25 & 14.10 & 21.6 & 19.86 & 4.55 \\ 
        \hline
        \midrule
       DGR \cite{choy2020deep} $\uparrow$ &  2.40 &  7.48 &  91.30 &  89.76 & 1.36\\ 
        \hline
        D3Feat$^*$ \cite{bai2020d3feat} & 2.57& 8.16 & 89.79 & 87.40 & 0.14 \\ 
        \hline
        SpinNet$^*$ \cite{ao2021spinnet} & 1.93 & 6.24& 93.74 & 92.07 & 2.84 \\ 
        \hline
       PointDSC \cite{bai2021pointdsc} & 2.06 & 6.55 & 93.28 & 89.35 & 0.09 \\ 
        \hline
       RoReg$^*$ \cite{wang2023roreg} $\downarrow$ &  1.84 & 6.28 &  93.70 &  91.60 & 2226 \\ 
        \hline
       Ours & $\mathbf{1.67}$ &  $\mathbf{5.68}$ & $\mathbf{94.60}$ & $\mathbf{94.35}$ & 0.12 \\ 
        \bottomrule
    \end{tabular}
\end{adjustbox}
\vspace{-0.2em}
\end{table}
\subsection{Outdoor Scenes Registration}
The input point cloud from the KITTI sequences \cite{geiger2012we} is downsampled using a voxel size of 30 cm to generate 1024 sparse points for the experiments.
\begin{table}[!th]
\centering
    \caption{Registration methods for evaluation on the KITTI dataset \cite{geiger2012we} involve testing the hand-crafted FPHF descriptor \cite{rusu2009fast} in conjunction with different learning strategies. The symbol $^*$ indicates the lack of support for replacing the feature descriptor, as per the original implementations on KITTI.}
    \label{tab:KITTI}
\begin{adjustbox}{width=0.82\textwidth}    
     \begin{tabular}{c c c c c c }
        \toprule 
        & RE($^{\circ}$) $\downarrow$ & TE(cm) $\downarrow$ & RR(\%) $\uparrow$ & F1(\%) $\uparrow$  & Time(s) $\downarrow$ \\
        \hline        
       RANSAC-1k \cite{fischler1981random} & 2.51 & 38.23 & 11.89 & 14.13 & 0.20\\
       \hline
       RANSAC-10k & 1.90 & 37.17 & 48.65 & 42.35 & 1.23\\
       \hline
       RANSAC-100k & 1.32 & 25.88 & 74.37 & 73.13 & 13.7\\
       \hline
       RANSAC-100k + refine & 1.28 & 18.42 & 77.20 & 74.07 & 15.65 \\
       \hline
        Go-ICP \cite{yang2015go} & 5.62 & 42.15 & 9.63 & 12.93 & 802\\
        \hline
        Super4PCS \cite{mellado2014super} & 4.83 & 32.27 & 21.04 & 23.72 & 6.29\\
        \hline
        \midrule
       FCGF-Reg$^*$ \cite{choy2019fully} $\downarrow$ & 1.95 &  18.51 &  70.86 &  68.90 & $\mathbf{0.09}$ \\
        \hline
       DGR \cite{choy2020deep} & 1.45 & 14.6 &  76.62 & 73.84 & 0.86 \\
        \hline
        D3Feat$^*$ \cite{bai2020d3feat} & 2.07 & 18.92 & 70.06 & 65.31& 0.23\\
        \hline
        SpinNet$^*$ \cite{ao2021spinnet} & 1.08 & 10.75 & 82.83 & 80.91 & 3.46 \\
        \hline
       PointDSC$^*$ \cite{bai2021pointdsc} & 1.63 & 12.31 &  74.41 & 70.08 & 0.31 \\
        \hline
       Ours & $\mathbf{0.92}$ & $\mathbf{8.74}$ & $\mathbf{83.83}$ & $\mathbf{85.09}$ & 0.14  \\
        \bottomrule
    \end{tabular}
\end{adjustbox}
\end{table}
We set the registration thresholds at 60cm for Translation Error and $5^{\circ}$ for Rotation Error. To measure Registration Recall (RR), we establish a threshold $\tau$ of 60cm. \cref{tab:KITTI} presents the quantitative results for comparison. Our proposed model demonstrates plausible performance compared to other methods, exhibiting minimal rotation and translation errors, and achieving the highest registration recall rate of $94.60\%$ when compared to RoReg, the second-best model. However, RoReg has a remarkable weakness in its real-time performance, registering in over 30 minutes. Additionally, we visually display the registration sample outcomes of our proposed model on 3DMatch and KITTI below. For a more comprehensive visual comparison to baselines, please refer to the supplementary section.\\
\begin{figure}[!th]
  \centering
  \includegraphics[width=0.94\linewidth]{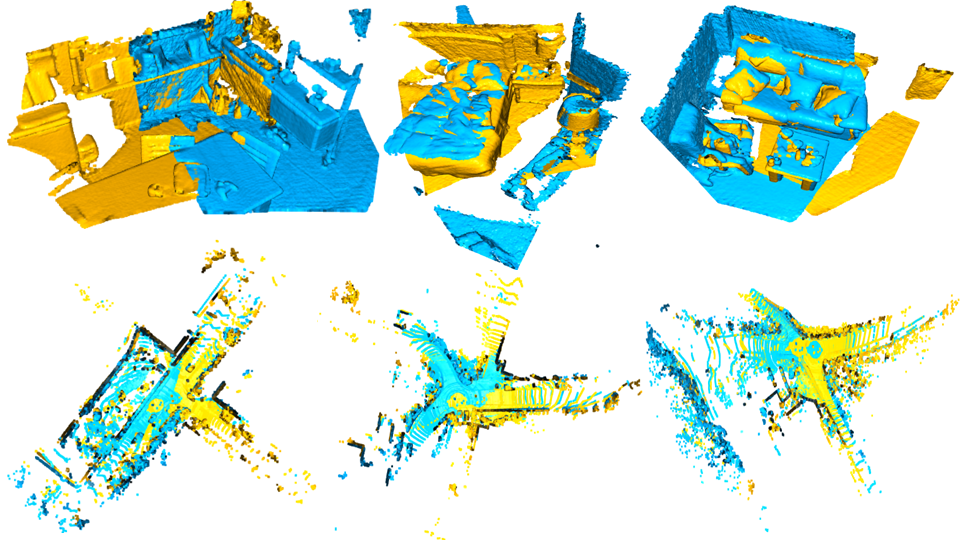}
  \caption{The visual registration results of the proposed model on 3DMatch \cite{zeng20173dmatch} and KITTI \cite{geiger2012we} are illustrated in the registration samples. Points from the target frame are represented in blue, whereas points converted from the source frame to the target frame by the predicted transform are visualized in yellow.}
 \label{fig:example}
 \vspace{-2.0em}
\end{figure}
To verify the proposed model performance under the different numbers of sampled input points, we also implement the sparse tests in \cref{tab:sample-pts}, by comparing with FCGF registration, D3Feat with or without PointDSC combination, and SpinNet. Our model has a consistent performance on 3DMatch over the comparison methods. In addition, using more points can boost the proposed model registration performance with a marginal gain. Considering a good trade-off between accuracy and computational efficiency, 1024 is chosen as the input point number in our final implemented model, because there is only a small performance gap within $1\%$ compared to the 2048 and 4096 number of points.
\begin{table}[!th]
\vspace{-1.2em}
\centering
    \caption{RR results on 3DMatch with a different number of sampled points.}
    \vspace{-0.8em}
    \label{tab:sample-pts}
\begin{adjustbox}{width=0.80\textwidth}    
     \begin{tabular}{c c c c c c c}
        \toprule 
        \textbf{\#Sampled Points} & $\mathbf{4096}$ \quad & $\mathbf{2048}$  \quad & $\mathbf{1024}$ \quad & $\hphantom{}\mathbf{512}$ \quad & $\hphantom{}\mathbf{256}$ \quad & $\textbf{Average}$ \\
        \hline        
       FCGF-Reg \cite{fischler1981random} & 91.7 & 90.3 & 89.5 & 85.7 & 80.5 & 87.5\\
       \hline
       D3Feat \cite{bai2020d3feat} & 91.9 & 90.4 & 89.8 & 86.0 & 82.5 & 88.1\\    
       \hline
       D3Feat \cite{bai2020d3feat}+PointDSC \cite{bai2021pointdsc} & 92.1 & 92.5 & 90.8 & 87.4 & 83.6 & 89.3\\
       \hline
       SpinNet \cite{ao2021spinnet} & 93.8 & 93.6 & 93.7 & 89.5 & 85.7 & 91.3\\
       \hline
       Ours & $\mathbf{95.3}$ & $\mathbf{94.8}$ & $\mathbf{94.6}$ & $\mathbf{91.3}$ & $\mathbf{88.5}$ & $\mathbf{92.9}$\\
        \bottomrule
    \end{tabular}
\end{adjustbox}
\vspace{-2.4em}
\end{table}
\subsection{Ablation Study}
Firstly, we present a table (\cref{tab:ablation}) displaying different configurations and combinations of module blocks in the proposed model.
\begin{table}[!th]
\vspace{-1.2em}
\centering
    \caption{An ablation study was conducted on our model design, using 3DMatch \cite{zeng20173dmatch} as the test dataset. The study involved comparing various descriptor combinations, exploring different combinations of layers for equi-feature learning, analyzing different methods of graph construction, and examining the impact of regularizers.}
    \label{tab:ablation}
\begin{adjustbox}{width=0.96\textwidth}    
     \begin{tabular}{l c c c c c c }
        \toprule 
       & & RE($^{\circ}$) $\downarrow$ & TE(cm) $\downarrow$ & RR(\%) $\uparrow$ & F1(\%) $\uparrow$  & Time(s) $\downarrow$ \\
       \hline
       \textbf{1.} & FCGF Descriptor + Ours & $\mathbf{1.62}$ & 6.24 & 93.87 & 94.28 & 0.15\\
       \hline
       \textbf{2.} & FPHF Descriptor + Ours & 1.83 & 6.49 & 83.62 & 73.06 & 0.12 \\
        \hline
        \hline
        \textbf{3.} & w/o Feature Descriptor Layers & 10.26 & 9.02 & 61.39 & 60.04 & 0.08\\
        \hline
        \textbf{4.} & w/o Equi-graph Layers & 9.64 & 8.37 & 62.45 & 60.03 & $\mathbf{0.06}$\\
        \hline
        \textbf{5.} & Replacing by normal GCNN & 8.32 & 5.94 & 68.52 & 67.54 & 0.10\\
        \hline
        \textbf{6.} & w/o LRFT layers & 2.76 & 6.47 & 83.09 & 81.78 & 0.09\\
        \hline
        \hline
        \textbf{7.} & KNN Graph Construction & 1.72 & $\mathbf{5.31 }$ & 92.37 & 93.74 & 0.16\\
        \hline
        \textbf{8.} & Ball Query Graph Construction & 1.67 & 5.68 & 94.60 & 94.35 & 0.12 \\
        \hline
        \hline
        \textbf{9.} & w/o Rank Regularizer & 6.41 & 7.92 & 76.45 & 78.09 & 0.10 \\
        \hline
        \textbf{10.} & w/o Sub-matrix Rank Verification & 2.58 & 6.93 & 87.76 & 88.36 & 0.07\\
        \hline
        \hline
       \textbf{11.} & Descriptor Layers + GCNN & 8.32 & 5.94 & 68.52 & 67.54 & 0.10\\
       \hline
       \textbf{12.} & w/o Descriptor Layers + Equi-GCNN & 10.26 & 9.02 & 61.39 & 60.04 & 0.08\\
       \hline
       \textbf{13.} & SpinNet + Equi-GCNN & 2.93 & 5.97 & 82.16 & 83.74 & 3.62 \\
       \hline
       \textbf{14.} & Ours & 1.67 & 5.68 & $\mathbf{94.60}$ & 9$\mathbf{4.35}$ & 0.12 \\
        \bottomrule
    \end{tabular}
\end{adjustbox}
\vspace{-2.0em}
\end{table}
Table analysis reveals significant accuracy enhancements with our model's learning-based descriptor over pre-trained FCGF and FPHF descriptors, as shown in rows 1 and 2, across multiple metrics. The descriptor learning layers (row 3) and equivariant graph CNN layers (row 4) are key to performance improvements. Replacing the equivariant with standard graph CNN layers (row 5) impairs rotation convergence, while omitting the LRFT module (row 6) marginally reduces performance. Ball query graph initialization (rows 8) outperforms KNN search (row 7) in efficiency and real-time outcomes on 3DMatch. Additionally, integrating rank regularizers (row 9) and sub-matrix rank verification (row 10) enhances model performance. Although equivariant layers introduce a slight computational delay (around 60ms), the overall performance gain is significant. Lastly, experiments of row 12-13 show that directly applying E-GCNN layers on the input or replacing the descriptor layers with equivariant structures such as SpinNet does not yield performance on par with our proposed method. Additionally, replacing E-GCNN layers with normal GCNN (row 11) results in significant performance degradation. These findings highlight the tailored effectiveness of our equi-approach.
\begin{figure}[!th]
    \vspace{-2.2em}
  \centering
\includegraphics[width=0.94\linewidth]{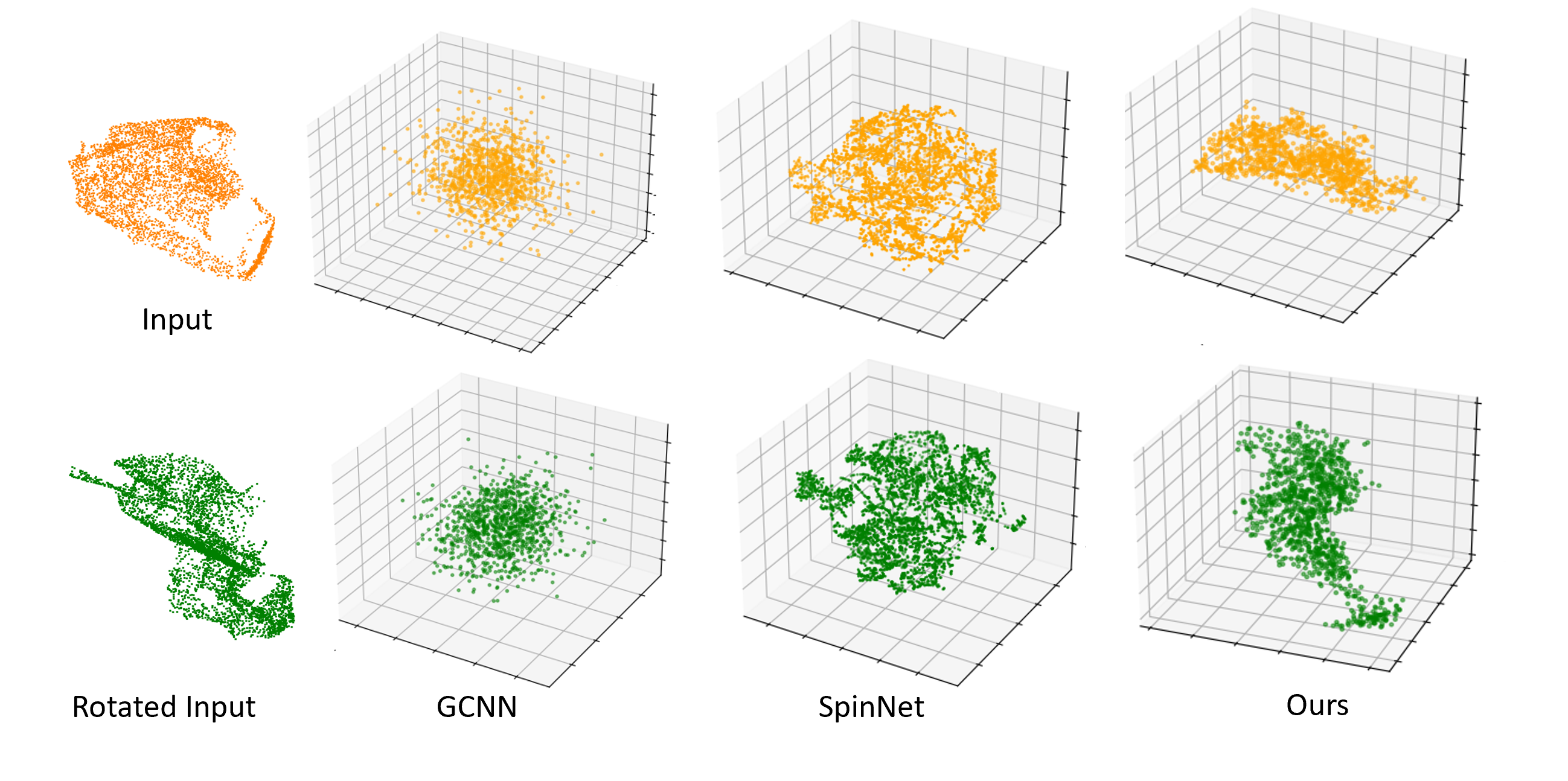}
  \caption{The t-SNE comparisons of equi-features outputs.}
 \label{fig:cmps-equi-feats}
 \vspace{-1.2em}
\end{figure}

For a more intuitive understanding of our model capability of learning equi-features, we provide t-SNE plots below for different encoder network outputs, showing that our Equi-Graph CNN produces features that are equivariant to rotated input. In comparison, features extracted by the conventional Graph CNN (GCNN) and SpinNet ($\mathbf{SO}(2)$ equivariant) as baselines do not exhibit rotation equivariance. To further elucidate the visual relationship between the rank of the feature similarity score matrix $\hat{\Vec{S}}$ and point correspondences with or without equivariant features, please refer to the supplementary section.

\begin{figure}[!ht]   
\vspace{-2.2em}
\begin{subfigure}{0.48\textwidth}
    \centering
    \vspace{-0.4em}
  \includegraphics[width=\textwidth, height=4.24cm]{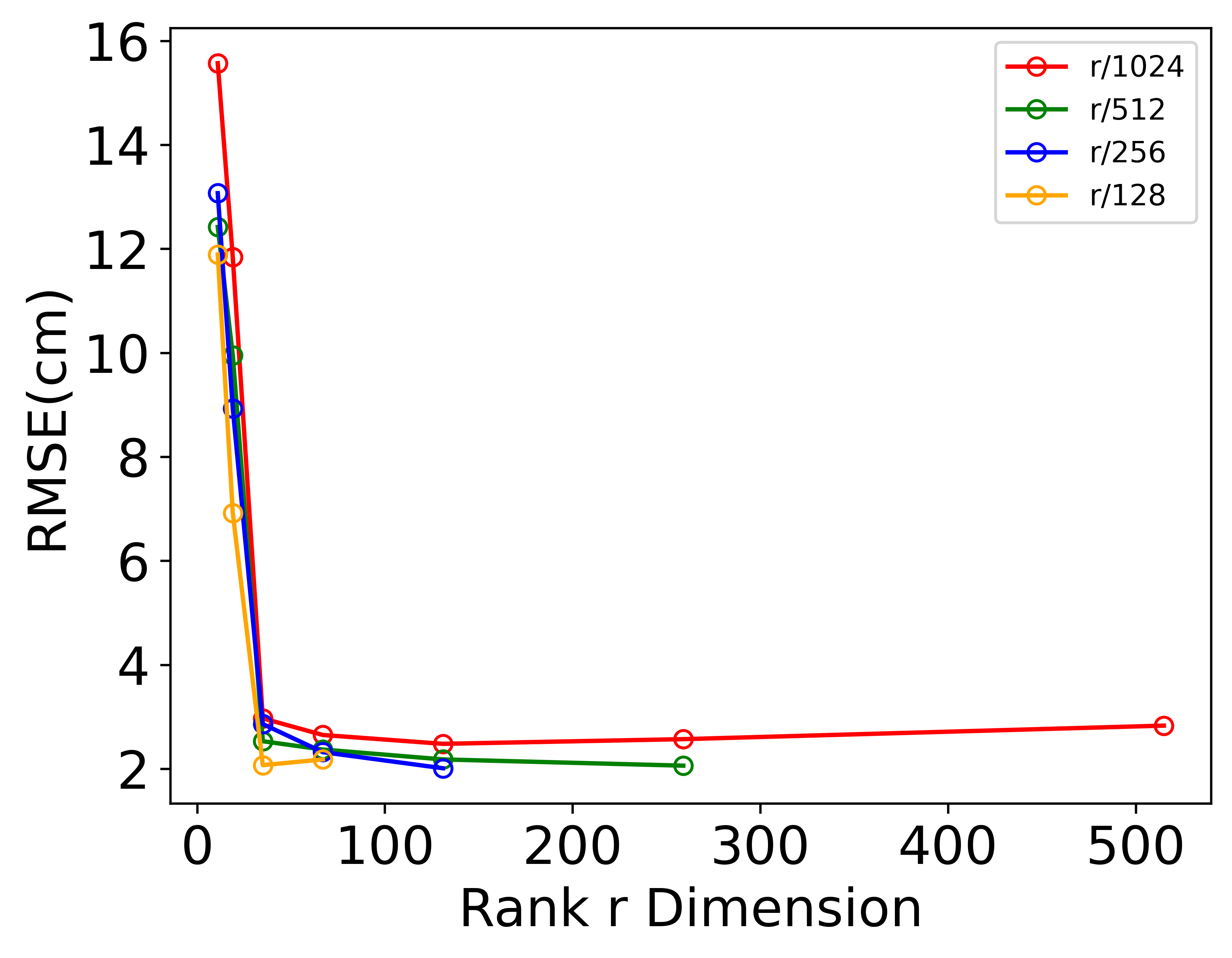}
    \caption{RMSE results under the different rank $r$ configurations, with four output dimension choices.}
    \label{fig:rank-ratio}
\end{subfigure}
\hfill
\begin{subfigure}{0.48\textwidth}
    \centering
    \includegraphics[width=\textwidth, height=4.8cm]{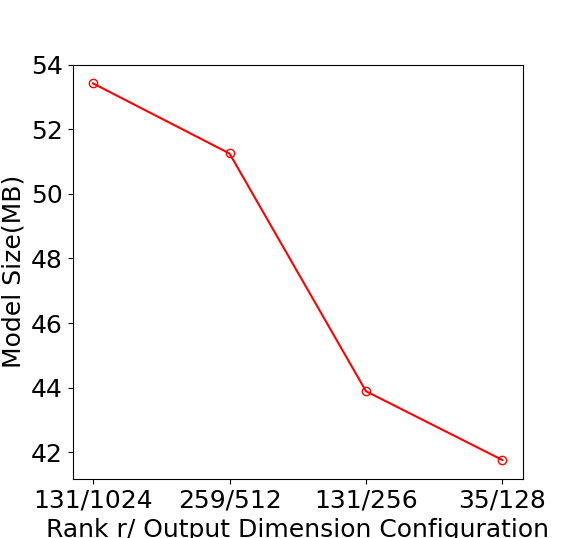}
    \caption{Model size under the four best-chosen rank/output dimension configurations.}
    \label{fig:mdoel-size-config}
\end{subfigure}
\caption{The left is the RMSE results plot of four output feature sizes of the LRFT module under the various rank dimensions, and the model size of the best $r$/output dimension configuration of each curve in the left plot is presented at right.}
\label{fig:rank-configs}
\vspace{-2.2em}
\end{figure}

Moreover, we provide the parameter configuration comparison of LRFT layers as well. It can be observed that varying the LRFT output feature number shows that increasing the middle-rank dimension enhances accuracy, yet beyond 200 ranks, error climbs a bit as depicted by the red curve. The relationship among the input feature number $N$, rank $r$, and the output number $N'$ is defined as $r \ll N' < N$, indicating that $N'$ should be at least twice the size of $r$, as evidenced by the termination position of curves in \cref{fig:rank-configs}. Additionally, RMSE as function of model size is plotted with various ($r/N'$) LRFT configuration, depicted by the individual colored curve on the left. 

\section{Conclusion}
We introduce an end-to-end model that leverages pre-trained feature descriptors or learns directly from raw scan points across two frames, incorporating equivariance embedding through graph layers, Low-Rank Feature Transformation and similarity score computation. Validation in both indoor and outdoor datasets confirms the superior performance of our proposed model. Ablation studies further substantiate the model design. Notably, the model's latency demonstrates its potential applicability in visual odometry. Future work could explore generalizing this framework to be input-order permutation invariant through graph attention layers or pooling, potentially integrating additional sensor modalities to address dynamic challenges.

\bibliographystyle{splncs04}
\bibliography{egbib}
\end{document}


\title{Supplementary Materials for \\ Equi-GSPR: Equivariant SE(3) Graph Network Model for Sparse Point Cloud Registration} 

\titlerunning{Equi-GSPR}

\author{Xueyang Kang\inst{1, 2, 3}\orcidlink{0000-0001-7159-676X} \and
Zhaoliang Luan\inst{2, 4}\orcidlink{0000-0002-9345-5455} 
\and
Kourosh Khoshelham\inst{3}\orcidlink{0000-0001-6639-1727}
\and
\\
Bing Wang\inst{2}\orcidlink{0000-0003-0977-0426}\thanks{Corresponding author}}

\authorrunning{X. Kang, Z. Luan, K. Khoshelham and B. Wang$^{\star}$}

\institute{Faculty of Electrical Engineering, KU Leuven \and Spatial Intelligence Group, The Hong Kong Polytechnic University \and Faculty of Engineering and IT, The University of Melbourne \and IoTUS Lab, Queen Mary University of London
\\
\email{alex.kang@kuleuven.com, z.luan@qmul.ac.uk, k.khoshelham@unimelb.edu.au, bingwang@polyu.edu.hk}
}
\maketitle

 \section{Equivariant Graph Network Model}
We provide a detailed view of the proposed model structure for transforming the stacked $N \times (32+3)$ tensors from the sparsely sampled input points, including the respective feature shape dimensions as indicated below. Implementation code can be found here, \url{https://github.com/alexandor91/se3-equi-graph-registration}.
\begin{figure}[!th]
    \vspace{-0.0em}
  \centering
  \includegraphics[width=\linewidth]{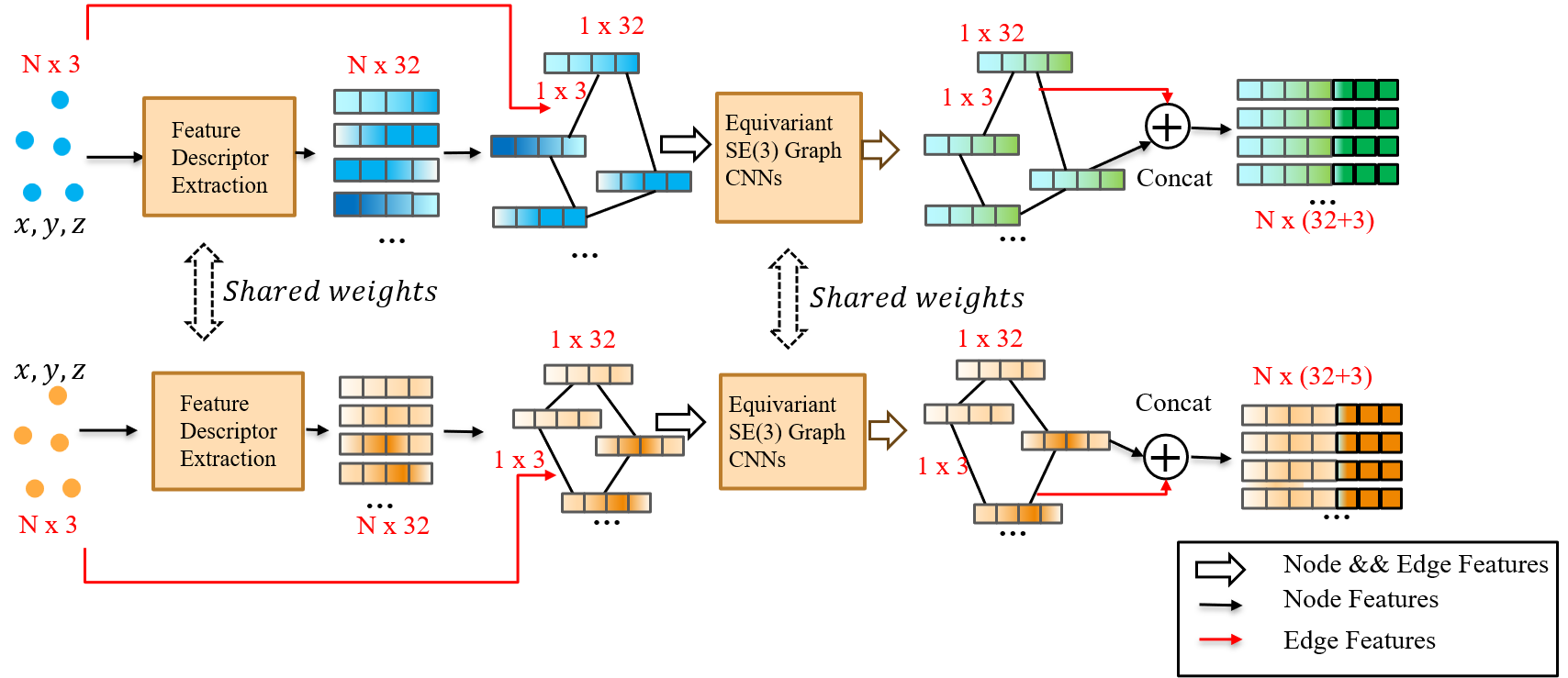}
    \caption{Initially, the process involves sparse input points from two frames, followed by the extraction of point-wise feature descriptors. Subsequently, a graph is constructed based on these descriptor features. The feature descriptor graphs then pass through individual equi-graph layers. The resulting graph node features are combined with the coordinate embedding to form tensors in shape of $N \times (32+3)$ for the decoder.}
    \label{fig:node-graph-convert}
 \vspace{-0.2em}
\end{figure}

The descriptor is generated point-wise from the sparsely sampled points in both the source and target frames. Subsequently, this descriptor, combined with the $\mathbf{SO}(3)$ edge coordinate embeddings discussed in the main body document, is used to create the feature graph. This graph is then inputted into graph convolution network layers. Following the equivariant graph layers, the resulting aggregated graph node features, combined with the average of neighbouring coordinate feature embeddings, are reorganized into a 2D tensor sized $N\times(32+3)$ before entering the Low-Rank Feature Transformation (LRFT) module.

\subsection{Graph Equivariance Proof}
Let's start by clarifying what equivariance means. Consider a graph $G = (V, E)$ with node features represented as $\mathbf{x}_v \in \mathbb{R}^{d}$ for each node $v \in V$, and a collection of transformations $\mathcal{G}$ that act on the graph. A Graph Neural Network (GNN) layer is considered equivariant to $\mathcal{G}$ if it meets the following criterion:
\begin{align}    
\mathbf{h}'_{v}  &= \rho_{\mathcal{G}}({(g \cdot \mathbf{x}_v, g \cdot \mathbf{e}_{u\rightarrow v}) \mid u \in \mathcal{N}(v)}) \\ &= g \cdot \rho_{\mathcal{G}}({\mathbf{x}_u, \mathbf{e}_{u\rightarrow v} \mid u \in \mathcal{N}(v)})
\end{align}

where $\mathbf{h}_v'$ represents the updated node feature for node $v$, $\rho_{\mathcal{G}}$ denotes the equivariant graph convolution operation, $\mathcal{N}(v)$ indicates the set of neighbours of node $v$, $\mathbf{e}_{u\rightarrow v}$ is the edge feature from node $u$ to node $v$, and $g \in \mathcal{G}$ represents a group transformation that operates on the node and edge features. To ensure equivariance, it is essential that the graph convolution operation $\rho_{\mathcal{G}}$ is formulated to preserve the group structure of $\mathcal{G}$. For instance, if $\mathcal{G}$ represents the permutation group that influences the node indices, $\rho_{\mathcal{G}}$ needs to remain unchanged when nodes are permuted.

Now, let's further discuss the concept of invariance. A Graph Neural Network (GNN) model is considered invariant to $\mathcal{G}$ if its final output, such as the graph prediction of a node, remains unchanged when subjected to group transformations $\mathcal{G}$. This can be mathematically represented as:
\begin{align}
\mathbf{y} = \rho_\text{inv}({\mathbf{h}_v \mid v \in V}) = \rho_\text{inv}({g \cdot \mathbf{h}_v \mid v \in V})    
\end{align}

where $\mathbf{y}$ is the final output, $\rho_\text{inv}$ is an invariant pooling operation (e.g., sum, max, or invariant multi-head attention), and ${\mathbf{h}_v \mid v \in V}$ are the node representations obtained after applying equivariant graph CNN layers.

To maintain invariance, the pooling operation $\rho_\text{inv}$ needs to be constructed in a way that remains unchanged when subjected to group transformations within $\mathcal{G}$. For instance, if $\mathcal{G}$ represents the permutation group, $\rho_\text{inv}$ should exhibit invariance towards permutations of nodes. By combining equivariance and invariance, a Graph Neural Network (GNN) can be formulated as follows,

Apply equivariant GNN layers to update node representations: 
\begin{align}
\mathbf{h}_v^{(l+1)} = \rho_{\mathcal{G}}^{(l)}({\mathbf{h}_u^{(l)}, \mathbf{e}_{u\rightarrow v} \mid u \in \mathcal{N}(v)})     
\end{align}

    Apply an invariant pooling operation to obtain the final output: $$\mathbf{y} = \rho_\text{inv}({\mathbf{h}_v^{(L)} \mid v \in V})$$

where $l$ is the layer index, and $L$ is the total number of GNN layers. 

This framework enables Graph Neural Networks (GNNs) to utilize group symmetries existing in the data, enhancing the efficiency and robustness of the learning process. The accurate configurations of the equivariant graph convolution operations denoted as $\rho_{\mathcal{G}}^{(l)}$ and the invariant pooling operation denoted as $\rho_\text{inv}$ are determined by the selected group $\mathcal{G}$ and the architecture of the GNN, here we use the sum and pooling operation to learn the invariance.

\subsection{Matrix Multiplication Rank Theorem Proof}
\textbf{Theorem 1.} Let $\Vec{A}$ be an $N \times r$ matrix, and $\Vec{B}$ be an $r \times N'$ matrix. We want to prove that:
\begin{equation}
\text{Rank}(\Vec{AB}) \leq \min(\text{Rank}(\Vec{A}), \text{Rank}(\Vec{B}))
\label{eq:RankAB}
\end{equation}
The proof relies on the following concepts, the rank of a matrix is equal to the dimension of its column space:

\textbf{Proof:} Let $\mathbf{A}$ be an $N \times r$ matrix and $\mathbf{B}$ be an $r \times N'$ matrix. The rank of a matrix is the maximum number of linearly independent columns (or rows) in the matrix. Let us denote the columns of $\mathbf{A}$ as $[\mathbf{a}_1, \mathbf{a}_2, \ldots, \mathbf{a}_r]$, and the rows of $\mathbf{B}$ as $[\mathbf{b}_1, \mathbf{b}_2, \ldots, \mathbf{b}_{r}]^T$. The results of $\mathbf{AB}$ are dot products of the columns of $\mathbf{A}$ and the rows of $\mathbf{B}$. Therefore, the maximum number of linearly independent columns in $\mathbf{AB}$ is bounded by the minimum of the number of linearly independent columns in matrix $\mathbf{A}$ or independent rows of $\mathbf{B}$.

To be noted, When the ratio of inliers to outliers in the feature correspondence is low, it can cause the feature similarity score matrix $\hat{\Vec{S}}$ in main body of paper to become rank-deficient with rank $r \ll 35$. This can result in difficulty for the training loss to converge due to uncertainty in the feature space. To tackle this issue, we adopt an iterative approach to seek a viable full-rank solution $r'$ that is smaller than $r$. This approach aims to minimize the overall training loss by ensuring that the full-rank condition of the submatrix is satisfied. The minimum value for the rank $r'$ is set at 16; any rank lower than this threshold may lead to a significant increase in registration errors through our experiments, consequently causing the registration process to fail.

\section{t-SNE Visualization of Equivariant Feature} 
\begin{figure}[!th]
    \vspace{-0.0em}
  \centering
  \includegraphics[width=0.68\linewidth]{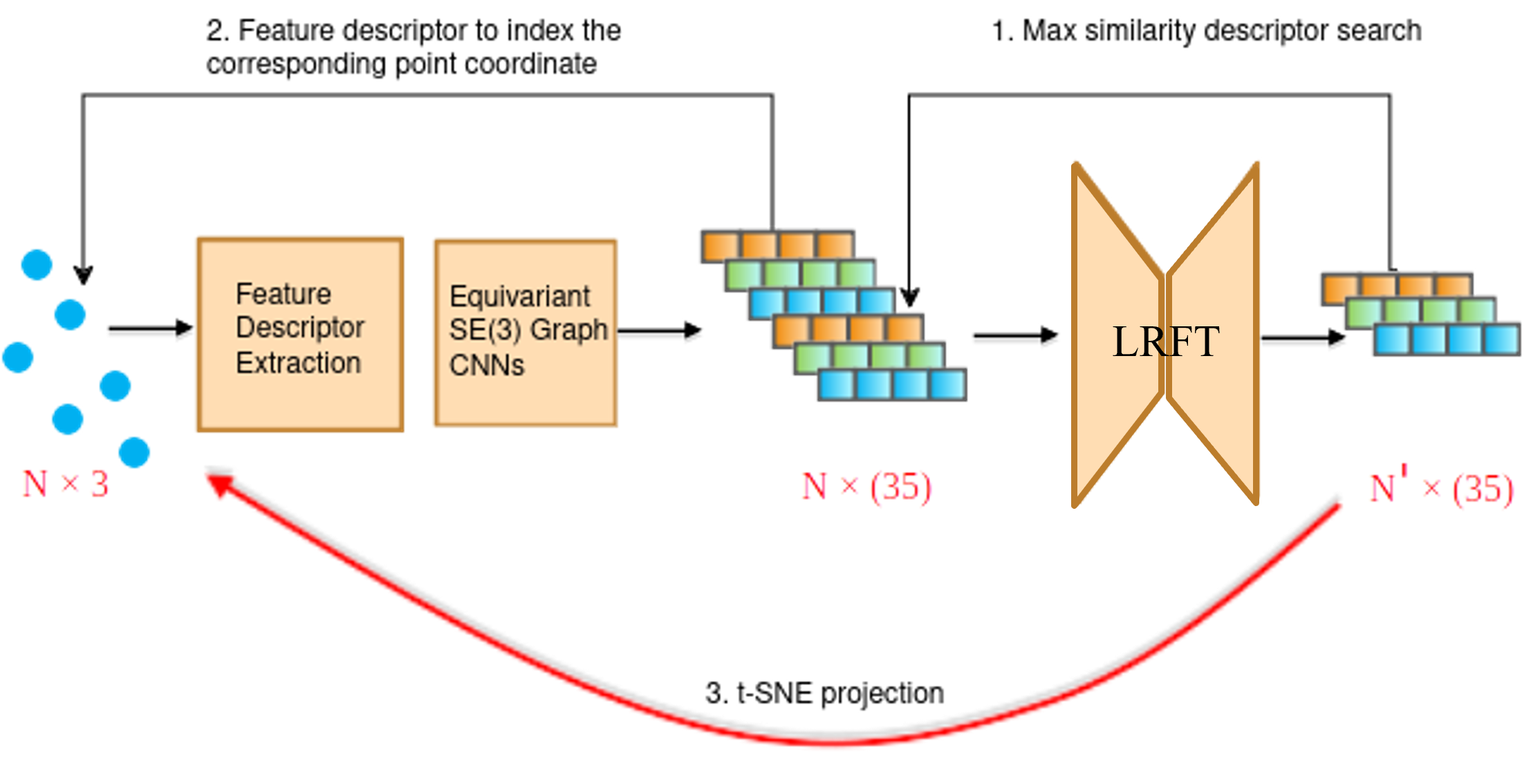}
    \caption{The pipeline uses t-SNE \cite{van2008visualizing} to map the sparse feature after the Low-Rank Feature Transformation (LRFT) module into the color map, then superimposed with the input points for visualization.}
    \label{fig:tsne-viz-pipe}
 \vspace{-1.2em}
\end{figure}

As illustrated in \cref{fig:tsne-viz-pipe}, The process includes mapping the graph feature through Low-Rank constrained MLP layers onto the input point coordinates.

\begin{figure}[!th]
    \vspace{-0.0em}
  \centering
  \includegraphics[width=\linewidth]{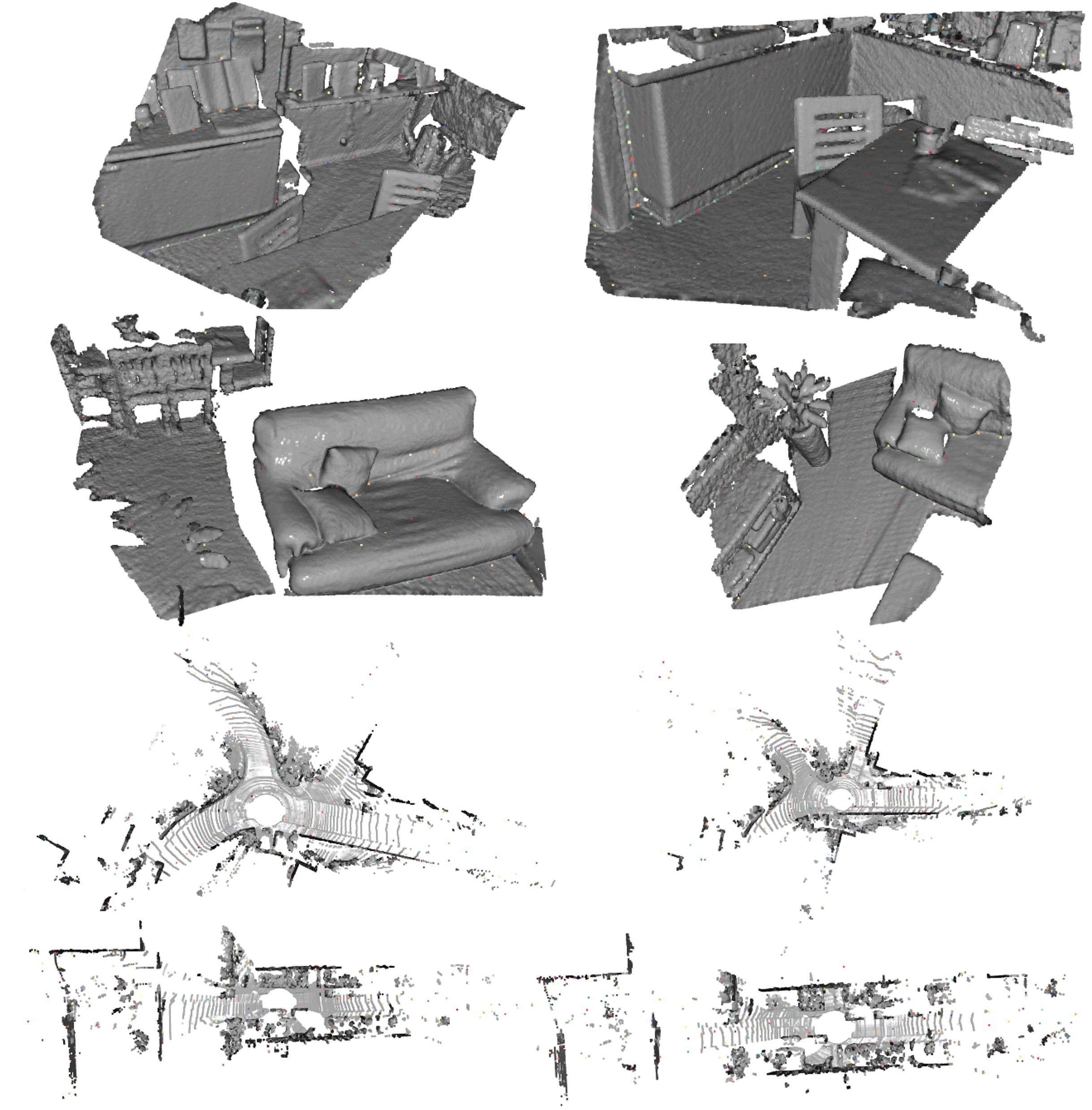}
    \caption{The left and right sides represent the source and target frame outcomes, respectively. Colored points (zoomed-in for better view) indicate mapped t-SNE feature values, while grey meshes depict raw input scans for visualization. The feature mapping pipeline employs t-SNE \cite{van2008visualizing} to map the output descriptors of post-Low-Rank Feature Transformation into a scalar color map. These are then superimposed onto respective input points from source and target frames to facilitate visualization.}
    \label{fig:tsne-points}
 \vspace{-2.0em}
\end{figure}

Initially, a feature similarity search is conducted by computing the dot product between feature vectors arranged along the row dimension, both pre- and post-LRFT, resulting in matrix shapes of $N\times 35$ and $N'\times 35$ respectively. The similarity matrix is then obtained through the Kronecker product, yielding an $N \times N'$ matrix. An $\argmax$ operation is applied to each row to identify the most similar pre-LRFT feature descriptor. The resulting index retrieves the corresponding input point coordinate, preserving feature descriptor association at the point cloud level. Notably, equi-graph layers maintain the original number and sequence of input points. These mapping stages establish a link between post-LRFT feature vectors and input points. Finally, 35-dimensional feature vectors are transformed into scalar color values using t-SNE \cite{van2008visualizing} and superimposed onto input point coordinates for visualization. \cref{fig:tsne-points} displays the descriptor feature output from the pipeline illustrated in \cref{fig:tsne-viz-pipe}. These features correspond to scalar values associated with input point coordinates of the initial input scan.

The mapped features are uniformly distributed throughout the scan, with notable concentrations along edges and corners (as evident on furniture surfaces in the first and second rows of \cref{fig:tsne-points}) (zoomed-in for better view). This visualization demonstrates that the equivariant features, post-equivariant graph layers, effectively represent distinctive geometric features in the input scan points. Furthermore, comparing the left and right column results reveals similar color and positional distributions of feature points between source and target frames, indicating a favorable correspondence distribution.

\section{Experiment Results}
We present additional visual comparison results of our model against baseline models, focusing on the top three quantitative performers from the main paper. For 3DMatch, the best baseline models include \textbf{SpinNet} \cite{ao2021spinnet}, \textbf{RoReg} \cite{wang2023roreg}, and \textbf{PointDSC} \cite{bai2021pointdsc}. For the KITTI dataset, we compare our model with \textbf{DGR} \cite{choy2020deep}, \textbf{SpinNet} \cite{ao2021spinnet}, and \textbf{PointDSC} \cite{bai2021pointdsc}, which represent the top three baseline models.

\begin{figure}[!th]
    \vspace{-0.0em}
  \centering
  \hspace{-0.8cm}\includegraphics[width=\linewidth]{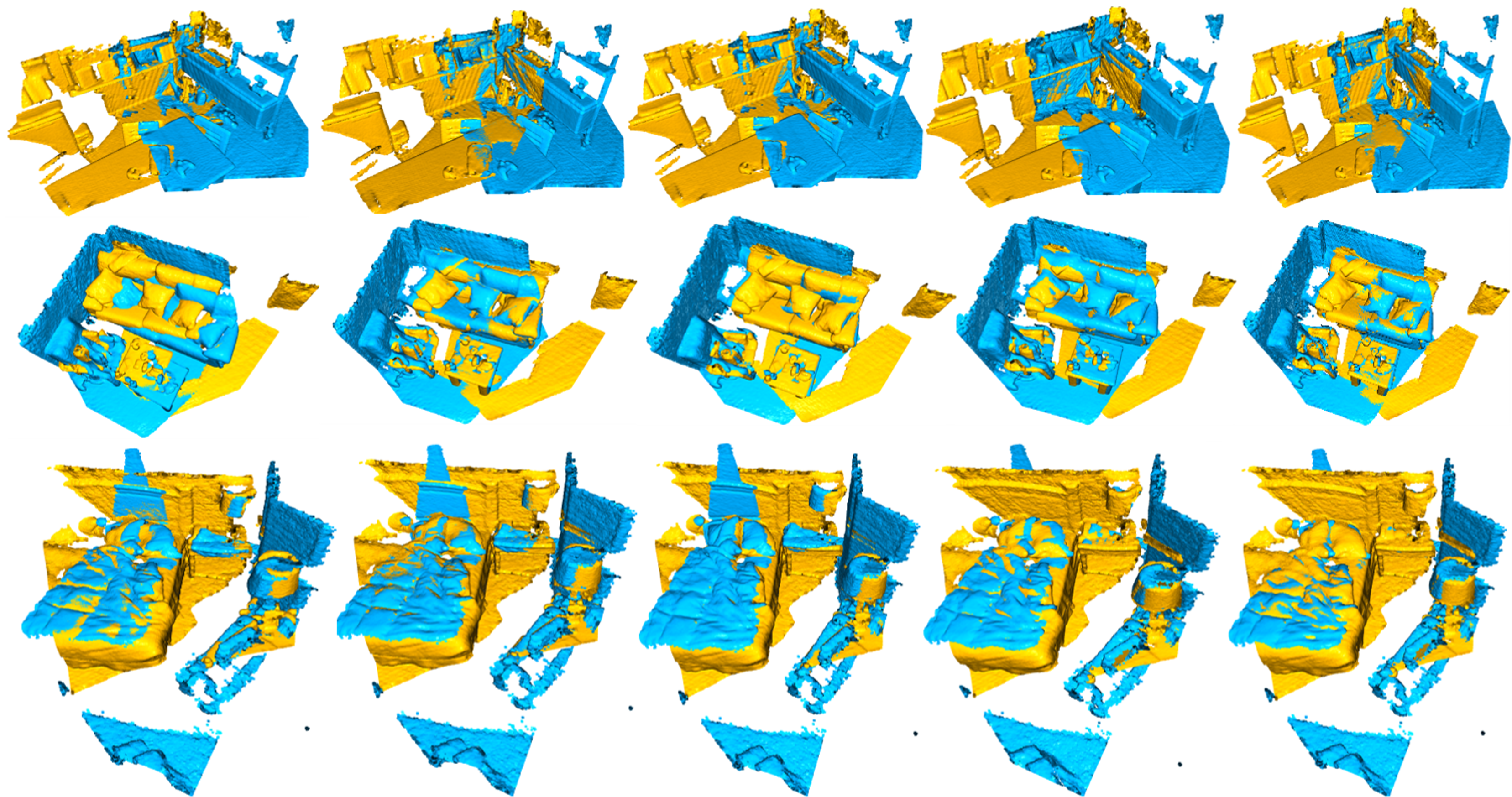}
    \qquad \qquad \qquad \qquad \qquad \qquad \qquad \textbf{PointDSC} \cite{bai2021pointdsc}\quad \quad \textbf{RoReg} \cite{wang2023roreg} \qquad \textbf{SpinNet} \cite{ao2021spinnet} \qquad \quad \textbf{Ours} \qquad \quad \textbf{Ground Truth}
    \caption{Visual comparisons on 3DMatch, the three models with top performance in the main paper are presented. Points from the target frame are represented in blue, whereas points converted from the source frame by the predicted transform are depicted in yellow.}
    \label{fig:cmps-3dmatch}
\end{figure}

Our proposed model demonstrates robust registration capabilities and high accuracy across diverse scenarios, contrasting with some learning models that exhibit performance degradation when transitioning from indoor to outdoor. This discrepancy is particularly evident in PointDSC's performance gap between 3DMatch (1st column of \cref{fig:cmps-3dmatch}) and KITTI (2nd column), where registration failure occurs in the initial KITTI case due to significant rotation errors.

\begin{figure}[!th]
    \vspace{-0.0em}
  \centering
  \hspace{-0.8cm} \includegraphics[width=\linewidth]{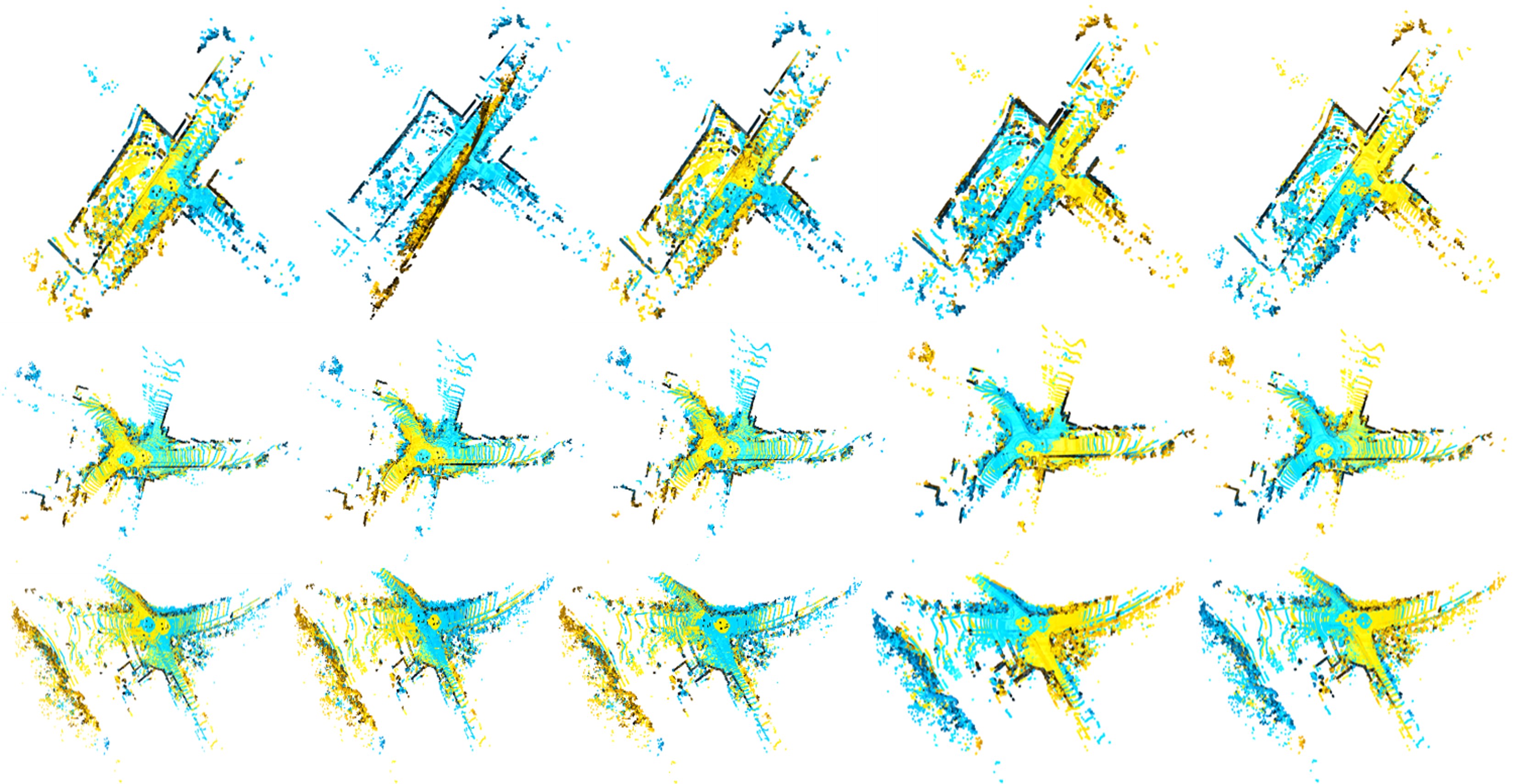}
  \qquad \qquad \qquad \qquad \qquad \textbf{DGR}\cite{choy2020deep} \quad \textbf{PointDSC}\cite{bai2021pointdsc} \quad \textbf{SpinNet} \cite{ao2021spinnet} \qquad\quad \textbf{Ours} \qquad \quad \textbf{Ground Truth}
    \caption{Visual comparison results on KITTI, the three models with top performance in the main paper are selected for visualization. Points from the target frame are colored in blue, whereas points converted from the source frame by the predicted transform are illustrated in yellow.}
    \label{fig:cmps=kitti}
\end{figure}

\subsection{Generalization on Unseen Datasets}
To demonstrate the robust generalization capabilities of our graph-based representation, we conduct a generalization test by directly evaluating the 3DMatch pre-trained model on the KITTI dataset, as detailed in \cref{tab:genral-test}. Each evaluation approach was repeated 10 times to report average performance and standard deviation of errors. Notably, our model consistently achieves a high registration recall rate of $82.31\%$, accompanied by minimal rotation and translation errors.
\begin{table}[!th]
\vspace{-0.4em}
\centering
    \caption{All the models are pre-trained on 3DMatch, and tested directly on KITTI.}
    \label{tab:genral-test}
\begin{adjustbox}{width=0.55\textwidth}    
     \begin{tabular}{c c c c c c }
        \toprule 
        &  \multicolumn{2}{c}{RE($^{\circ}$) $\downarrow$} & \multicolumn{2}{c}{TE(cm) $\downarrow$}
         & \multirow{2}{*}{RR(\%) $\uparrow$}  \\
        &  AVG & STD & AVG & STD & \\
        \hline
       FCGF \cite{fischler1981random} & 1.61 & 1.51 & 27.1 & 5.58 & 24.19 \\
       \hline
       D3Feat(rand) \cite{bai2020d3feat} & 1.44 & 1.35 & 31.6 & 10.1 & 36.76\\
       SpinNet \cite{ao2021spinnet} & 0.98 & $\mathbf{0.63}$ & 15.6 & 1.89 & 81.44 \\
        \hline
       Ours & $\mathbf{0.86}$ & 0.68 & $\mathbf{10.7}$ & $\mathbf{1.23}$ & $\mathbf{82.31}$ \\
        \bottomrule
    \end{tabular}
\end{adjustbox}
\vspace{-1.2em}
\end{table}

%
\section{Ablation Study}
\begin{figure}[!th]
\vspace{-0.2em}
  \centering
  \includegraphics[width=0.86\textwidth]{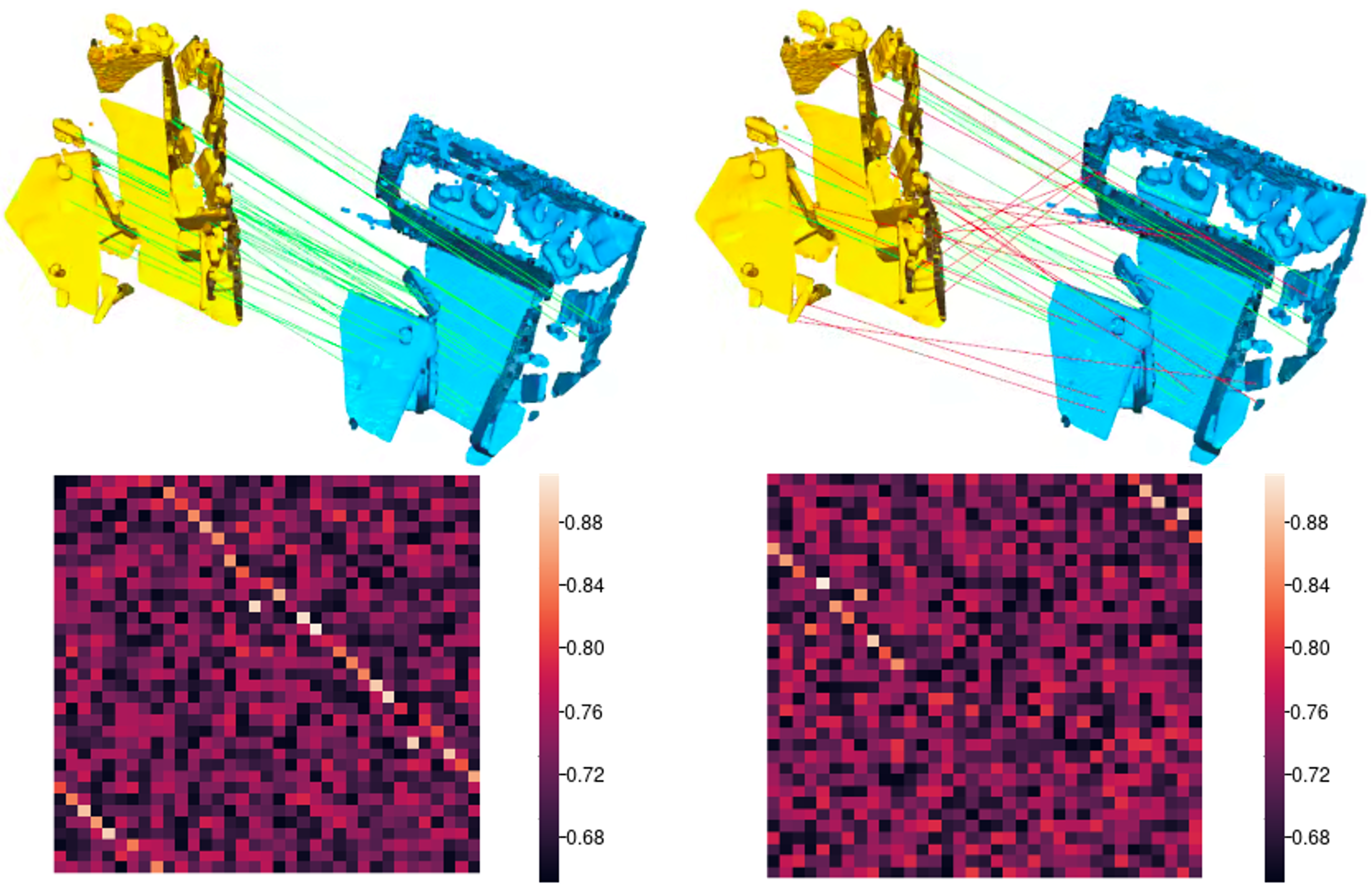}
  \caption{This figure illustrates the comparison between correspondence results with and w/o equivariant features in graph layers horizontally, and the relationship vertically between the feature similarity score matrix and point correspondence. The top row displays the visualization of point correspondence linked to feature similarity below.}
 \label{fig:feature-sim-corres}
\vspace{-1.6em}
\end{figure}
To demonstrate the relationship between the rank of the feature similarity score matrix $\hat{\Vec{S}}$ and point correspondence, we specifically examine the model with normal graph CNN layers or equi-graph CNN layers. This analysis includes cases both with and without incorporating equivariance into graph layers to facilitate a comparative visualization of equivariance impact for registration. We extract the top 35 similarity score values from each case to identify pairwise features. These pairwise features are then linked with the feature descriptor before the LRFT module by selecting the maximum descriptor similarity in each row of similarity score matrix. Subsequently, this process allows us to retrieve the input point coordinates. By following these procedures, the feature pair after LRFT module can be correlated with the input point for visualization. The bottom row of \cref{fig:feature-sim-corres} indicates that the adoption of equivariant features significantly enhances feature distinctiveness. This enhancement facilitates the generation of valid, unique rank values within the similarity score matrix, as illustrated on the left side matrix rank. In contrast, the application of non-equivariant features tends to increase ambiguity in match score computations. Moreover, a binary indicator $ \omega_i$ for visual assessment of point correspondences requires the comparison of distance errors against the inlier threshold $\tau$ as below,
\begin{align}
    \vspace{-4.0em}
    \omega_i &= \mathds{1}[ \| \hat{\Vec{R}} \Vec{x}_i + \hat{\Vec{t}} - \Vec{y}_j\| < \tau],
    \label{eq:corrsp-node-label}
    \vspace{-4.0em}
\end{align}
The label one is depicted as a green line while the zero is represented by the red line in the top row of \cref{fig:feature-sim-corres}. 

\begin{figure}[!th]
    \vspace{-2.4em}
  \centering
  \includegraphics[width=0.76\linewidth]{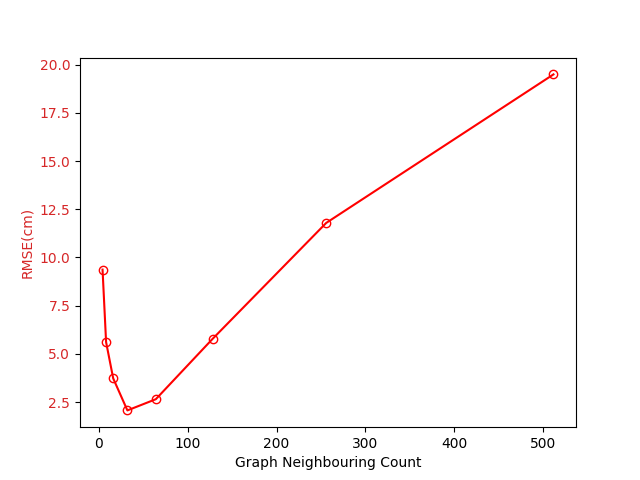}
    \caption{RMSE plot as a function of  graph neighbouring feature node count.}
    \label{fig:nn-count}
 \vspace{-2.0em}
\end{figure}

Finally, we evaluated the impact of neighboring node count on feature graph initialization and accuracy performance, as measured by the RMSE metric (please refer to \cref{fig:nn-count}).

Performance improves as the number of neighboring nodes used for graph creation increases to 24. However, a significant performance deterioration occurs when the node count exceeds 200. This decline suggests that an excessive number of neighboring feature nodes can cause overflow of information for the whole graph feature learning, dispersing attention during feature aggregation and consequently reducing performance due to ambiguous neighboring features. While the plot displays a graph node count limit of 512, the actual limit is 1024, which unfortunately triggers out-of-memory issues during model computations in our hardware settings.

\clearpage